\documentclass[journal]{IEEEtran}
\usepackage{graphicx}
\usepackage{amsmath}
\usepackage{amsfonts}
\usepackage{amssymb}
\usepackage{subfigure}
\usepackage{color}
\usepackage{cite}
\usepackage{bm}
\usepackage{mathrsfs}
\usepackage{makecell}
\usepackage{multirow}
\usepackage{balance}

\newtheorem{theorem}{Theorem}
\newtheorem{remark}{Remark}

\newtheorem{example}{Example}

\ifCLASSINFOpdf
\else
\fi
\begin{document}
\graphicspath{{figures/}}
%
\title{Simultaneous Position and Orientation Planning of Nonholonomic Multi-Robot Systems: A Dynamic Vector Field Approach}
%
%
%

\author{Xiaodong~He and
        Zhongkui~Li,~\IEEEmembership{Senior Member,~IEEE}
\thanks{This work was supported by the National Natural Science Foundation of China under grants 61973006, T2121002, and by Beijing Natural Science Foundation, China under grant JQ20025. Corresponding author: Zhongkui Li. }
\thanks{The authors are with the State Key Laboratory for Turbulence and Complex Systems, Department of Mechanics and Engineering Science, College of Engineering, Peking University, Beijing 100871, China (e-mail: hxdupc@pku.edu.cn; zhongkli@pku.edu.cn)}
}

\maketitle

\begin{abstract}
This paper considers the simultaneous position and orientation planning of nonholonomic multi-robot systems. Different from common researches which only focus on final position constraints, we model the nonholonomic mobile robot as a rigid body and introduce the orientation as well as position constraints for the robot's final states. In other words, robots should not only reach the specified positions, but also point to the desired orientations simultaneously. The challenge of this problem lies in the underactuation of full-state motion planning, since three states need to be planned by mere two control inputs. To this end, we propose a dynamic vector field (DVF) based on the rigid body modelling. Specifically, the dynamics of the robot orientation are brought into the vector field, implying that the vector field is not static on the 2-D plane anymore, but a dynamic one varying with the attitude angle. Hence, each robot can move along the integral curve of the DVF to arrive at the desired position, and in the meantime, the attitude angle can converge to the specified value following the orientation dynamics. Subsequently, by designing a circular vector field under the framework of the DVF, we further study the obstacle avoidance and mutual-robot-collision avoidance in the motion planning. Finally, numerical simulation examples are provided to verify the effectiveness of the proposed methodology.
\end{abstract}


%
\IEEEpeerreviewmaketitle

\section{Introduction}

Motion planning is a fundamental problem in the field of robotics and control, which aims at finding paths or trajectories to guide mobile robots moving from initial conditions to their respective destinations, while avoiding collisions with obstacles and other robots. One of the most typical robotic systems is the nonholonomic mobile robot, which is also referred to as the unicycle-type vehicle. Although neglected by a number of persons, two facts regarding the nonholonomic mobile robot should be mentioned above all. One is that the theoretical model of such a kind of robot is essentially a rigid body rather than a point of mass \cite{Bloch2015Nonholonomic}, since the robot states include both position and orientation (or attitude). Different from a 3-degree-of-freedom (DOF) rigid body moving freely on a plane, such a robot has no lateral velocity due to the nonholonomic constraint, which naturally brings about the other fact that the nonholonomic mobile robot is indeed an underactuated system \cite{Bullo2005Geometric}. To be more specific, the robot has three DOFs (two translation DOFs and one rotation DOF), while possessing only two control inputs (one linear velocity and one angular velocity). These two facts make the motion planning for nonholonomic mobile robots more challenging and demanding.

Although a variety of methodologies have been proposed for motion planning, such as roadmap \cite{Kavraki1996Probabilistic,Bhattacharya2008Roadmap,Lehner2018Repetition}, cell decomposition \cite{Cai2009Information,Zhang2008Efficient,Cowlagi2012Multiresolution}, sampling-based algorithm \cite{Karaman2011Sampling,Jaillet2010Sampling,Oh2021Chance-Constrained}, they cannot be applied to nonholonomic mobile robots resulting from omitting the kinematics or only considering the model of the point of mass. Regarding the nonholonomic model, some researchers employ the optimization-based methods \cite{Hussein2008Optimal,Hausler2016Energy,Zhao2021Pareto,Bloch2021Dynamic,Li2021Efficient,Cichella2021Optimal,Zhao2022Scalable,Li2021Optimal}. For example, Bloch et al. in \cite{Bloch2021Dynamic} formulate the problem as dynamic interpolation on Riemannian manifolds and provide the necessary conditions for optimality. Li et al. in \cite{Li2021Efficient} propose a prioritized optimization method so as to compute the planning results efficiently. In \cite{Cichella2021Optimal}, Cichella et al. utilize the Bernstein polynomials to transform the motion planning into a discrete optimization approximately. There is no denying that optimization methods have the advantage of easily handling unicycle models, because the nonholonomic constraints can always be included as the optimization constraints. Nevertheless, the feasibility of such optimization problems are unable to be guaranteed, or they suffer from extremely heavy computational burden. Another drawback is that the control inputs derived from optimization are open-loop, relying only on time, thereby are not robust to disturbances.

Given above shortages, researchers are motivated to investigate feedback motion planning algorithms for mobile robots. A typical closed-loop methodology is the velocity vector field, where a velocity vector related to the state is defined at every point in the configuration space and the integral curve of the vector field converges to the goal point. The most common vector field is defined by the gradient of a potential function, also referred to as the potential field \cite{Ge2002Dynamic,Huang2009Velocity,Valbuena2012Hybrid,Karagoz2014Coordinated,Kovacs2016A_novel,Tian2021An_Overall}, but one of its inherent limitations is the possible local minina other than the desired state \cite{Koren1991Potential}. Particular forms of potential function overcome such a drawback, that is, harmonic function \cite{Kim1992Real-time,Garrido2010Garrido,Masoud2012Motion} and navigation function \cite{Rimon1992Exact,Loizou2008Navigation,Li2019Navigation}. However, the former has demanding computational complexity related to PDEs, while the latter is difficult to implement since a lower bound to ensure no local minima is unknown in advance.

As a matter of fact, the velocity vector field does not necessarily have to be given by a potential gradient. Instead, it can be defined directly over the configuration space. To the best of our knowledge, few works focus on motion planning via non-gradient-based vector fields, except two seminal papers by Lindemann et al. \cite{Lindemann2009Simple} and Panagou \cite{Panagou2017A_Distributed}. In \cite{Lindemann2009Simple}, the environment with obstacles is decomposed into convex polytopes, in which simple local vector fields are defined and smoothly blended to form a global vector field convergent to the desired point. The author in \cite{Panagou2017A_Distributed} proposes a family of 2-D analytic vector fields, which exhibit different patterns (such as attractive or repulsive) by choosing the value of an parameter, so that the overall vector field can finally be obtained by a suitable blending design.

Regardless of open-loop or feedback algorithms, the aforementioned results rarely consider a significant yet implicit fact typically occurring in real-world scenarios. Actually, apart from the position constraints, the orientation of a robot is also usually required to reach a desired direction at the terminal time. For instance, in the multi-robot surveillance mission, the final orientation of each robot should point to a certain direction so as to obtain the largest  overall surveillance area. Similarly, the attitudes of missiles are generally specified in the terminal guidance so as to realize a better performance of coordinated attack. Thus, these practical demands strongly motivate us that it is necessary to incorporate the orientation constraints into the motion planning. Additionally, such a motivation is further strengthened from a theoretical point of view. As mentioned above, the model of a nonholonomic mobile robot is a rigid body. Then, serving as a state, the attitude angle should also be specified to be a desired value, similar to the position, rather than being left randomly, which can be referred to as full-state motion planning. Since the nonholonomic mobile robot is an underactuated system, the biggest challenge lies in how to achieve full-state planning by fewer control inputs.

Motivated by above discussions, in this paper we consider the simultaneous position and orientation planning of nonholonomic multi-robot systems by designing a non-gradient-based velocity vector field. To begin with, the mobile robot is modelled as a rigid body with nonholonomic constraints rather than a point of mass. More importantly, besides the desired position constraint, we take into account the orientation constraint at the final time instant as well. Next, different from common static vector fields on a 2-D plane, we propose a novel dynamic vector field (DVF) in the sense that the dynamics of the attitude angle are introduced into the vector field. This implies that the velocity direction at a certain point is decided by not only the position but also the orientation of robot that passes such a point. Thus, by moving along the integral curve of the DVF, the robot can reach the specified position at the terminal time, and meanwhile the attitude angle can converge to the desired value following the orientation dynamics. Subsequently, the control inputs or velocities are designed based on the DVF, where the feedback of an extra angle in the body-fixed frame is brought in as an additional angular velocity, with the result that the robot orientation can be tuned along the direction of the DVF to deal with the lack of lateral velocity. Moreover, under the frame work of the DVF, the problems of obstacle avoidance and mutual-robot-collision avoidance are studied by proposing a circular vector field, where robots can move along the tangential directions of the round obstacles so as to evade collisions. Note that such two kinds of collision avoidance problems are both able to be solved by one circular vector field, which greatly simplifies the design of planned trajectories.

Apart from the dynamical characteristics, the following merits also distinguish our method from the existing motion planning results which also utilize non-gradient-based vector fields. As opposed to \cite{Lindemann2009Simple}, the proposed dynamic vector field is unnecessary to be derived based on a cell decomposition of the whole environment, then advanced high-level discrete motion planning is not required in this paper. In contrast to \cite{Panagou2017A_Distributed}, the dynamic vector field is global over the state space in the sense that the initial and final configurations can both be chosen arbitrarily, while the vector field in \cite{Panagou2017A_Distributed} has a separatrix (or mirror line) where the integral curves can become divergent possibly.

The organization of this paper is given below. Section~\ref{sec_preli} provides mathematical preliminaries and formulates the motion planning problem. Section~\ref{sec_DVF} proposes the dynamic vector field approach, under which the problems of obstacle avoidance and collision avoidance are solved by designing a circular vector field in Section~\ref{sec_ob_avoid} and Section~\ref{sec_co_avoid}, respectively. Section~\ref{sec_sim} gives several numerical simulation examples to verify the effectiveness of the proposed vector fields. Finally, conclusions end the whole paper in Section~\ref{sec_con}.

\section{Preliminaries and Problem Statement}\label{sec_preli}

Several commonly-used notations are defined in advance. The identity matrix in $\mathbb{R}^n$ is denoted by $\bm{I}_n$. The base vectors of $\mathbb{R}^3$ are denoted by $\bm{e}_1,\bm{e}_2,\bm{e}_3$. The symbol $\bm{0}_{m\times n}$ represents a matrix in $\mathbb{R}^{m\times n}$ with all zero components. The Euclidean norm of a vector is denoted by $\|\cdot\|$. Moreover, variables denoting vectors and matrices are written in bold, while those denoting scalars are not.

As mentioned in the Introduction, the model of the nonholonomic mobile robot is a rigid body. Thus, before providing the nonholonomic model, we firstly consider a fully actuated planar rigid body moving in a 2-D Euclidean space. Let $\bm{\mathcal{F}}_{\mathcal{E}}$ denote the earth-fixed frame, and let $\bm{\mathcal{F}}_{\mathcal{B}}$ represent the body-fixed frame, which is attached to the center of mass of the rigid body. The position of the rigid body in $\bm{\mathcal{F}}_{\mathcal{E}}$ is given by a vector $\bm{p}=[x\ \ y]^{\rm T}\in\mathbb{R}^2$, and the attitude is specified by a rotation matrix $\bm{R}\in\mathbb{R}^{2\times 2}$, which depicts the rotation of $\bm{\mathcal{F}}_{\mathcal{B}}$ relative to $\bm{\mathcal{F}}_{\mathcal{E}}$. Herein, the rotation matrix $\bm{R}$ can be parameterized by a scalar $\theta\in[-\pi,\pi]$, that is,
\begin{equation}\label{eq_rotation_matrix}
  \bm{R}=\begin{bmatrix}
      \cos\theta & -\sin\theta \\
      \sin\theta & \cos\theta
    \end{bmatrix},
\end{equation}
where $\theta$ is interpreted as the attitude angle of the rigid body. Let $\omega\in\mathbb{R}$ and $\bm{v}=[v^x\ v^y]^{\rm T}\in\mathbb{R}^2$ denote the rigid body's angular velocity and linear velocity in the body-fixed frame $\bm{\mathcal{F}}_{\mathcal{B}}$. Then, the kinematics of the fully actuated rigid body can be given by
\begin{subequations}\label{eq_kine_fully}
    \begin{align}
      \dot{x} &= v_x\cos\theta-v_y\sin\theta, \\
      \dot{y} &= v_x\sin\theta+v_y\cos\theta, \\
      \dot{\theta} &= \omega.
    \end{align}
\end{subequations}
Regarding nonholonomic mobile robots, due to no side slip of the wheels, the robot cannot move sideways. In other words, the velocity along the $Y_{\mathcal{B}}$-axis of the body-fixed frame $\bm{\mathcal{F}}_{\mathcal{B}}$ is always zero, that is, $v_y=0$. Such a constraint is named the nonholonomic constraint, since its reformulation in the earth-fixed frame $\bm{\mathcal{F}}_{\mathcal{E}}$ which is a differential equation given by
\begin{equation*}
  \dot{x}\sin\theta-\dot{y}\cos\theta=0
\end{equation*}
cannot be integrated to be an algebraic equation. Thus, according to (\ref{eq_kine_fully}), the kinematic model of a nonholonomic mobile robot degenerates to
\begin{subequations}\label{eq_kine_nonho}
    \begin{align}
      \dot{x} &= v_x\cos\theta, \\
      \dot{y} &= v_x\sin\theta, \\
      \dot{\theta} &= \omega.
    \end{align}
\end{subequations}

Given the fact that the nonholonomic mobile robot moves in an environment possibly populated with obstacles, we assume that each obstacle can be bounded by a circular region with radius $r_o>0$. Moreover, suppose that multiple nonholonomic mobile robots move in a common workspace simultaneously. Therefore, the potential collisions with obstacles and among robots should both be taken into account in the motion planning process. To acquire the information of obstacles and other robots positions, the robot $i$ ($i=1,\cdots,N$) is assumed to have a circular sensing range with radius $R_s$, which can be defined by
\begin{equation*}
  \mathcal{S}_i=\left\{\bm{q}\in\mathbb{R}^2\  \big| \ \|\bm{q}-\bm{p}_i\|\leq R_s \right\},
\end{equation*}
where $\bm{p}_i$ is the current position vector of the robot $i$. Then, once obstacles and other robots appear into the sensing region $\mathcal{S}_i$, the robot $i$ can obtain the position information of them.

The motion planning problem for multiple nonholonomic mobile robots can be formulated below.

\emph{Problem Statement:} Regarding $N$ nonholonomic mobile robots described by (\ref{eq_kine_nonho}), design linear velocity $v_{ix}$ and angular velocity $\omega_i$ for the robot $i$ ($i=1,\cdots,N$), such that each controlled trajectory of the robots, which starts from the initial condition $(x_{i0},y_{i0},\theta_{i0})$, can reach the specified destination $(x_{id},y_{id},\theta_{id})$, and meanwhile avoid the obstacles and mutual collisions with other robots.

\section{Dynamic Vector Fields}\label{sec_DVF}

In this section, we construct a non-gradient-based vector field to solve the motion planning problem for a nonholonomic mobile robot in an obstacle-free environment. With the loss of generality, we assume that the destination of the robot $(x_d,y_d,\theta_d)$ is chosen as $(0,0,0)$. For the sake of simplicity, we firstly rewrite the kinematics (\ref{eq_kine_fully}) and (\ref{eq_kine_nonho}) in a more compact form.

Define the following matrix
\begin{equation}\label{eq_confi_matrix}
  \bm{h}=\begin{bmatrix}
      \bm{R} & \bm{p} \\
      \bm{0}_{1\times2} & 1
    \end{bmatrix}=
    \begin{bmatrix}
      \cos\theta & -\sin\theta & x \\
      \sin\theta & \cos\theta  & y \\
      0 & 0 & 1
    \end{bmatrix},
\end{equation}
which is uniquely decided by the rotation matrix $\bm{R}$ and the position vector $\bm{p}$, and we call $\bm{h}$ the configuration of the rigid body. Similarly, the velocity can be formulated in a matrix as
\begin{equation}\label{eq_velo_matrix}
  \bm{\eta}=\begin{bmatrix}
              \hat{\bm{\omega}} & \bm{v} \\
              \bm{0}_{1\times2} & 0
            \end{bmatrix}=
            \begin{bmatrix}
              0 & -\omega & v_x \\
              \omega & 0 & v_y \\
              0 & 0 & 0
            \end{bmatrix},
\end{equation}
where the hat operator $\hat{\cdot}$ defines a map from a scalar to a skew symmetric matrix in $\mathbb{R}^{2\times2}$. Therefore, based on (\ref{eq_confi_matrix}) and (\ref{eq_velo_matrix}), the fully actuated kinematic model (\ref{eq_kine_fully}) can be redefined by
\begin{equation}\label{eq_kine_fully_m}
  \dot{\bm{h}}=\bm{h}\bm{\eta},
\end{equation}
where the configuration $\bm{h}$ and the velocity $\bm{\eta}$ serve as the state and control input, respectively. Correspondingly, the nonholonomic kinematic model can be rewritten as (\ref{eq_kine_fully_m}) with an additional nonholonomic constraint
\begin{equation*}
  \bm{e}_2^{\rm T}\bm{\eta}\bm{e}_3=0,
\end{equation*}
where $\bm{e}_i$ ($i=1,2,3$) are standard basis in $\mathbb{R}^3$.

To design the vector field, a transformation is introduced for the configuration $\bm{h}$. By utilizing the matrix logarithmic map proposed in \cite{Bullo2005Geometric,Bullo1995Proportional}, we obtain the logarithm of configuration $\bm{h}$ as follows
\begin{equation}\label{eq_Upsi_log_h_def}
  \bm{\Upsilon}=\log(\bm{h})=\begin{bmatrix}
                               \hat{\bm{\theta}} & \bm{\varphi}(x,y,\theta) \\
                               \bm{0}_{1\times 2} & 0
                             \end{bmatrix},
\end{equation}
where $\hat{\bm{\theta}}$ is a $2\times 2$ skew symmetric matrix with respect to $\theta$, similar to the form of $\hat{\bm{\omega}}$ in (\ref{eq_velo_matrix}), and the vector $\bm{\varphi}(x,y,\theta)$ is given by
\begin{equation}\label{eq_phi_def}
  \bm{\varphi}(x,y,\theta)=\frac{\theta}{2}
  \begin{bmatrix}
    \frac{1+\cos\theta}{\sin\theta} & 1 \\
    -1 & \frac{1+\cos\theta}{\sin\theta}
  \end{bmatrix}
   \begin{bmatrix}
     x \\
     y
   \end{bmatrix}
   \triangleq
   \begin{bmatrix}
     \varphi_1(x,y,\theta) \\
     \varphi_2(x,y,\theta)
   \end{bmatrix}.
\end{equation}
Once defining a matrix
\begin{equation}\label{eq_Xi_for_phi}
  \bm{\Xi}=\frac{\theta}{2}
  \begin{bmatrix}
    \frac{1+\cos\theta}{\sin\theta} & 1 \\
    -1 & \frac{1+\cos\theta}{\sin\theta}
  \end{bmatrix},
\end{equation}
the formula (\ref{eq_phi_def}) can be simplified as
\begin{equation}\label{eq_phi_def_simp}
  \bm{\varphi}(x,y,\theta)=\bm{\Xi}\bm{p}.
\end{equation}
where $\bm{p}$ is the position vector. Then, we refer to $\bm{\Upsilon}$ given in (\ref{eq_Upsi_log_h_def}) as the transformed configuration, which is uniquely defined by the attitude angle $\theta$ and position vector $\bm{p}$. According to \cite{Bullo1995Proportional}, under the transformed configuration $\bm{\Upsilon}$, the kinematics (\ref{eq_kine_fully_m}) can be reformulated as
\begin{equation}\label{eq_dot_log_h}
  \dot{\bm{\Upsilon}}=\bm{\mathcal{M}}(\bm{\Upsilon})\bm{\eta},
\end{equation}
where the transformed configuration $\bm{\Upsilon}$ is the state, the velocity $\bm{\eta}$ is the control input, and $\bm{\mathcal{M}}(\bm{\Upsilon})\in\mathbb{R}^{3\times 3}$ is a state-dependent matrix satisfying
\begin{equation}\label{eq_M_Upsi_property}
  \bm{\mathcal{M}}(\bm{\Upsilon})\bm{\Upsilon}=\bm{\Upsilon}.
\end{equation}
We recommend \cite{Bullo1995Proportional} for readers who are interested in more information about the kinematics (\ref{eq_dot_log_h}) and matrix $\bm{\mathcal{M}}(\bm{\Upsilon})$.

\begin{remark}
  As a matter of fact, the formulas (\ref{eq_confi_matrix})-(\ref{eq_M_Upsi_property}) are all originated from the geometric control theory of mechanical systems. Specifically, the configuration $\bm{h}$ is an element in the Lie group ${\rm SE}(2)$; the body velocity $\bm{\eta}$ is a twist in the Lie algebra $\mathfrak{se}(2)$; $\bm{\Upsilon}$ is called the exponential coordinate of $\bm{h}$, which also lies in the Lie algebra $\mathfrak{se}(2)$ but represents the configuration. Since the framework of geometric control is established on fully actuated rigid bodies, we introduce related preliminaries at the beginning of Section~\ref{sec_preli}, in advance of the nonholonomic mobile robots. Although we subsequently design the vector field with these variables in geometric control, for the sake of readability, the concepts such as Lie group and Lie algebra are not introduced in this paper. This is because we do not utilize any complicated properties from geometric control theory, and it is believed that the aforementioned modelling preliminaries are sufficient for the completeness and explicitness of this paper. The readers interested in geometric control are recommended to refer to \cite{Bullo2005Geometric,Murray1994A_Mathematical,Bloch2015Nonholonomic}.
\end{remark}

Now, we propose a vector field $\bm{\Gamma}_d:\mathbb{R}^2\times\mathbb{S}\to\mathbb{R}^2$, whose components $\Gamma_{d}^x$ and $\Gamma_{d}^y$ are given by
\begin{subequations}\label{eq_DVF}
  \begin{align}
    \Gamma_{d}^x &= -\varphi_1(x,y,\theta)\cos\theta+\varphi_2(x,y,\theta)\sin\theta, \\
    \Gamma_{d}^y &= -\varphi_1(x,y,\theta)\sin\theta-\varphi_2(x,y,\theta)\cos\theta,
  \end{align}
\end{subequations}
where $\theta$ serve as an internal parameter satisfying
\begin{equation}\label{eq_dot_theta_minus_theta}
  \dot{\theta}=-\theta.
\end{equation}

\begin{figure}[htp]
  \centering
  \subfigure[$\theta(0)=\frac{\pi}{2}$]{
    \label{fig_DVF_pi_2}
    \includegraphics[width=0.23\textwidth,trim=60 5 70 5,clip]{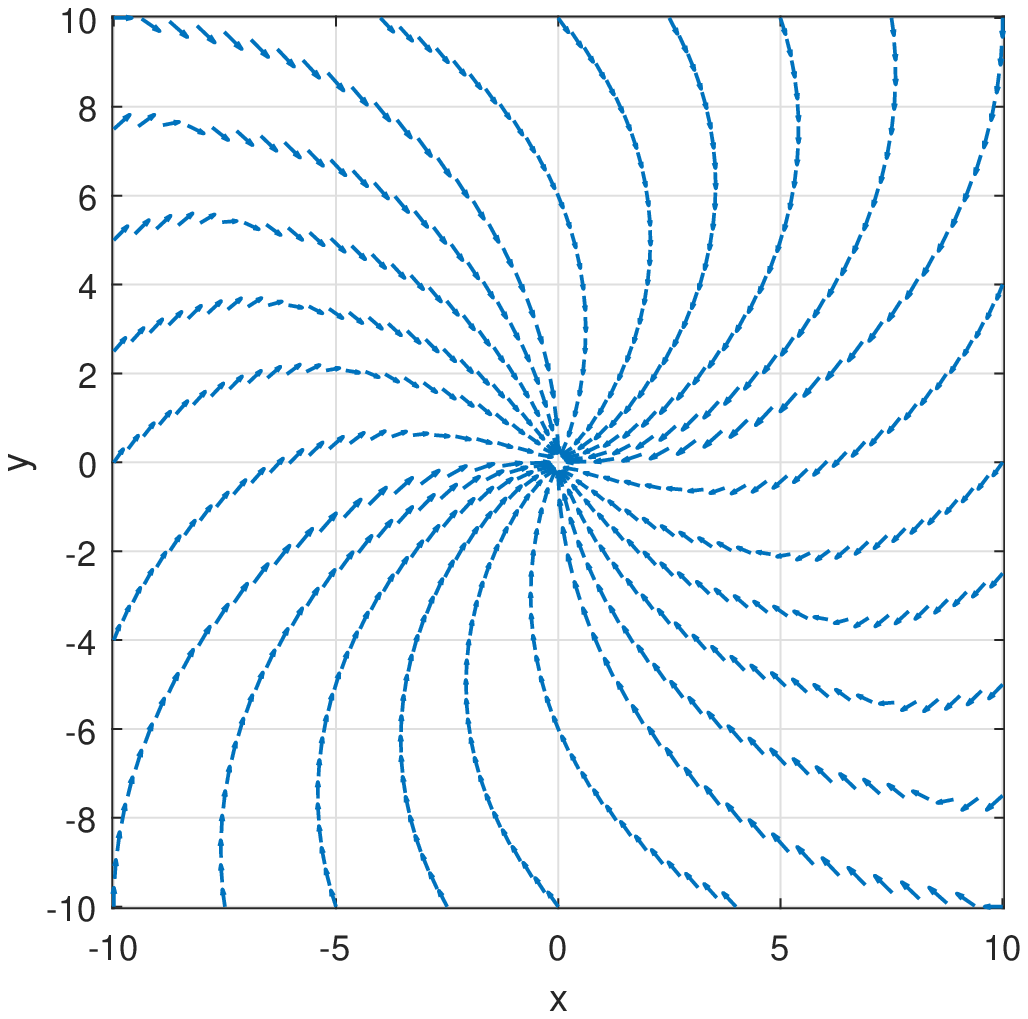}}
  \subfigure[$\theta(0)=-\frac{\pi}{2}$]{
    \label{fig_DVF_minus_pi_2}
    \includegraphics[width=0.23\textwidth,trim=60 5 70 5,clip]{pi_2_VF.eps}}
  \caption{Dynamic vector field $\bm{\Gamma}_d$ with different initial parameter $\theta$}
  \label{fig_DVF}
\end{figure}

\begin{remark}
  In contrast to common planner vector fields which are maps of $\mathbb{R}^2\to\mathbb{R}^2$, such as in \cite{Panagou2017A_Distributed,Kapitanyuk2018A_Guiding}, one more parameter $\theta\in\mathbb{S}$ is introduced to the vector field $\bm{\Gamma}_d$ in this paper. More importantly, $\theta$ has an explicit physical meaning, that is, the attitude angle of the robot. To bring $\theta$ into the vector field is naturally motivated by the fact that the nonholonomic mobile robot is essentially a kind of rigid bodies rather than a point of mass. Consequently, as a state of the robot, the attitude angle are not supposed to be set free at the destination. Instead, it should be guided to the specified value, like positions, after the motion planning procedure. Moreover, the designation of the final attitude angle has a more practical value in the sense that it specifies the initial velocity direction of the subsequent period of motion. Thus, we incorporate the attitude information into the vector field, leading to the result that $\bm{\Gamma}_d$ is decided by not only position but also orientation. Then, the vector field in $\mathbb{R}^2$ is ``dynamic" instead of a ``static" one. Herein, the ``dynamic" means $\bm{\Gamma}_d$ will vary with the initial value of $\theta$, which evolves following (\ref{eq_dot_theta_minus_theta}). Therefore, even if for the same one point, the vector direction is probably different due to the attitude angle of the robot. In light of this fact, we refer to $\bm{\Gamma}_d$ as a dynamic vector field. Fig.~\ref{fig_DVF} provides two dynamic vector fields with initial condition ${\theta(0)}=\frac{\pi}{2}$ and ${\theta(0)}=-\frac{\pi}{2}$, respectively.
 \end{remark}

 The convergence of $\bm{\Gamma}_d$ is provided in the following theorem.

\begin{theorem}\label{theo_DVF}
  The dynamic vector field $\bm{\Gamma}_d$ given in (\ref{eq_DVF}), (\ref{eq_dot_theta_minus_theta}) converges to $(x,y,\theta)=(0,0,0)$ asymptotically.
\end{theorem}

\begin{IEEEproof}
  Firstly, we prove that the transformed configuration $\bm{\Upsilon}$ converges to $\bm{0}$ asymptotically. Let $\dot{x}=\Gamma_{d}^x$ and $\dot{y}=\Gamma_{d}^y$. By comparing with the kinematics (\ref{eq_kine_fully}), we obtain that the velocity or the control input $\eta$ can be expressed as
  \begin{equation}\label{eq_eta_minus_Upsi}
    \bm{\eta}=-\begin{bmatrix}
            \hat{\bm{\omega}} & \bm{v} \\
            \bm{0}_{1\times 2} & 0
          \end{bmatrix}
          =-\begin{bmatrix}
            \hat{\bm{\theta}} & \bm{\varphi}(x,y,\theta) \\
            \bm{0}_{1\times 2} & 0
          \end{bmatrix}=-\bm{\Upsilon},
  \end{equation}
  where the condition (\ref{eq_dot_theta_minus_theta}) is utilized. Define a Lyapunov function $\Phi=\frac{1}{2}\langle \bm{\Upsilon},\bm{\Upsilon} \rangle$, where $\langle \cdot,\cdot \rangle$ represents the inner product. Taking the time derivative of $\Phi$ along the trajectory of (\ref{eq_dot_log_h}), we have
  \begin{equation}\label{eq_dot_Phi_1}
    \dot{\Phi}=\langle \bm{\Upsilon},\dot{\bm{\Upsilon}} \rangle = \langle \bm{\Upsilon},\bm{\mathcal{M}}(\bm{\Upsilon})\bm{\eta} \rangle.
  \end{equation}
  Substituting (\ref{eq_eta_minus_Upsi}) into (\ref{eq_dot_Phi_1}), there holds
  \begin{equation}\label{eq_dot_Phi_2}
    \dot{\Phi}=-\langle \bm{\Upsilon},\bm{\mathcal{M}}(\bm{\Upsilon})\bm{\Upsilon} \rangle=-\langle \bm{\Upsilon},\bm{\Upsilon} \rangle < 0, \quad \forall\bm{\Upsilon}\ne\bm{0},
  \end{equation}
  where the property (\ref{eq_M_Upsi_property}) is used. Thus, we obtain that the transformed configuration $\bm{\Upsilon}$ converges to $\bm{0}$ asymptotically, implying $\theta\to 0$ and $\bm{\varphi}(x,y,\theta)\to \bm{0}$ as $t\to\infty$.

  Next, we prove that the position vector $\bm{p}=[x\ \ y]^{\rm T}$ can asymptotically converge to $\bm{0}$. By using the L'Hospital's rule, we have
  \begin{equation*}
    \lim_{\theta\to 0}\frac{\theta}{2}\frac{1+\cos\theta}{\sin\theta} = \lim_{\theta\to 0}\frac{\theta}{\sin\theta} = \lim_{\theta\to 0}\frac{1}{\cos\theta}=1,
  \end{equation*}
  so that there holds
  \begin{equation*}
    \lim_{\theta\to 0}{\det}(\bm{\Xi})=1,
  \end{equation*}
  where ${\rm det}(\cdot)$ represents the determinant and the matrix $\bm{\Xi}$ is given in (\ref{eq_Xi_for_phi}). Therefore, based on the formulation in (\ref{eq_phi_def_simp}), we obtain that $\bm{\varphi}(x,y,\theta)\to \bm{0}$ if and only if $\bm{p}\to\bm{0}$.
\end{IEEEproof}

In the following, we extend the dynamic vector field $\bm{\Gamma}_d$ to an arbitrarily-specified final state $(x_d,y_d,\theta_d)$. Based on (\ref{eq_confi_matrix}), once there holds $(x,y,\theta) = (0,0,0)$, the corresponding configuration matrix $\bm{h}$ becomes a $3\times 3$ identity matrix $\bm{I}_3$. Thus, the dynamic vector field $\bm{\Gamma}_d$ given in (\ref{eq_DVF}), (\ref{eq_dot_theta_minus_theta}) can drive $\bm{h}$ to $\bm{I}_3$, indeed. Let $\bm{h}_d$ denote the corresponding configuration matrix of the final state $(x_d,y_d,\theta_d)$. Then, the problem now becomes how to define a dynamic vector field $\tilde{\bm{\Gamma}}_d$ which can drive the configuration $\bm{h}$ to $\bm{h}_d$. Motivated by the fact that $\bm{\Gamma}_d$ achieves $\bm{h}\to\bm{I}_3$,, we define a new configuration
\begin{equation}\label{eq_tilde_h}
  \tilde{\bm{h}}=\bm{h}_d^{-1}\bm{h},
\end{equation}
which contains the information of the desired final state $\bm{h}_d$. The parameterized description of $\tilde{\bm{h}}$, denoted by $(\tilde{x},\tilde{y},\tilde{\theta})$, can be given by $\tilde{x}=(x-x_d)\cos\theta_d+(y-y_d)\sin\theta_d$, $\tilde{y}=-(x-x_d)\sin\theta_d+(y-y_d)\cos\theta_d$, $\tilde{\theta}=\theta-\theta_d$. Based on above definitions, we present the following theorem.

\begin{theorem}
  For an arbitrarily-specified final state $(x_d,y_d,\theta_d)$, design a dynamic vector field $\tilde{\bm{\Gamma}}_d:\mathbb{R}^2\times\mathbb{S}\to\mathbb{R}^2$, whose components $\tilde{\Gamma}_{d}^x$ and $\tilde{\Gamma}_{d}^y$ are given by
  \begin{subequations}\label{eq_DVF_arbitr}
    \begin{align}
      \tilde{\Gamma}_{d}^x &= -\varphi_1(\tilde{x},\tilde{y},\tilde{\theta})\cos\tilde{\theta}+\varphi_2(\tilde{x},\tilde{y},\tilde{\theta})\sin\tilde{\theta}, \\
      \tilde{\Gamma}_{d}^y &= -\varphi_1(\tilde{x},\tilde{y},\tilde{\theta})\sin\tilde{\theta}-\varphi_2(\tilde{x},\tilde{y},\tilde{\theta})\cos\tilde{\theta},
    \end{align}
  \end{subequations}
  where $\tilde{\theta}$ satisfies $\dot{\tilde{\theta}}=-\tilde{\theta}$. Then, the dynamic vector field $\tilde{\bm{\Gamma}}_d$ converges to $(x,y,\theta)=(x_d,y_d,\theta_d)$ asymptotically.
\end{theorem}

\begin{IEEEproof}
  According to Theorem~\ref{theo_DVF}, the dynamic vector field $\tilde{\bm{\Gamma}}_d$ will converge to $(\tilde{x},\tilde{y},\tilde{\theta})=(0,0,0)$ asymptotically. That is to say, $\tilde{\bm{\Gamma}}_d$ can drive the configuration $\tilde{\bm{h}}$ to the identity matrix $\bm{I}_3$. Then, according to the definition of $\tilde{\bm{h}}$ in (\ref{eq_tilde_h}), we have $\bm{h}_d^{-1}\bm{h}\to\bm{I}_3$, i.e., $\bm{h}\to\bm{h}_d$, implying that $(x,y,\theta)\to(x_d,y_d,\theta_d)$.
\end{IEEEproof}

Having obtained the dynamic vector field $\tilde{\bm{\Gamma}}_d$, we can further design the control inputs $\omega$ and $v_x$. It should be noted that the dynamic vector field $\tilde{\bm{\Gamma}}_d$ in (\ref{eq_DVF_arbitr}), (\ref{eq_dot_theta_minus_theta}) is defined in the earth-fixed frame $\bm{\mathcal{F}}_{\mathcal{E}}$. In contrast, the control inputs are the velocities given in the body-fixed frame $\bm{\mathcal{F}}_{\mathcal{B}}$. Thus, for the purpose of a simple controller design, we transform the dynamic vector field $\tilde{\bm{\Gamma}}_d$ from the earth-fixed frame $\bm{\mathcal{F}}_{\mathcal{E}}$ to the body-fixed frame $\bm{\mathcal{F}}_{\mathcal{B}}$. Let $\tilde{\bm{\Gamma}}_{d}^{\mathcal{B}}$ denote the dynamic vector field in $\bm{\mathcal{F}}_{\mathcal{B}}$. Then, $\tilde{\bm{\Gamma}}_{d}^{\mathcal{B}}$ can be obtained by
\begin{equation}\label{eq_DVF_in_Fb}
  \tilde{\bm{\Gamma}}_{d}^{\mathcal{B}}=\bm{R}^{\rm T}\tilde{\bm{\Gamma}}_d\triangleq
  \begin{bmatrix}
    \tilde{\Gamma}_{d}^{\mathcal{B}x} \\
    \tilde{\Gamma}_{d}^{\mathcal{B}y}
  \end{bmatrix}.
\end{equation}
where $\bm{R}$ is the rotation matrix defined in (\ref{eq_rotation_matrix}). Thus, based on (\ref{eq_DVF_in_Fb}), it will be straightforward to design the motion planning controller of fully actuated rigid bodies, which can be obtained by making the body velocity $\bm{\eta}$ equivalent to $\tilde{\bm{\Gamma}}_{d}^{\mathcal{B}}$, that is
\begin{equation}\label{eq_fully_controller}
  v_x=k_v\tilde{\Gamma}_{d}^{\mathcal{B}x},\quad v_y=k_v\tilde{\Gamma}_{d}^{\mathcal{B}y},\quad \omega=-k_{\omega}\theta,
\end{equation}
where $k_v,k_{\omega}$ are positive scalars.

\begin{figure}[htp]
  \centering
  \includegraphics[width=0.25\textwidth]{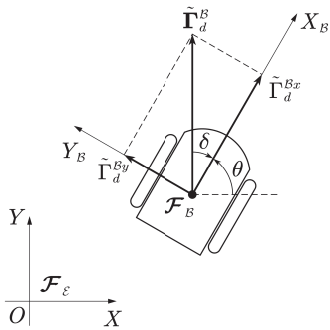}
  \caption{Orientation of the nonholonomic mobile robot and direction of the vector field $\bm{\Gamma}$}
  \label{fig_vehicle_attitude}
\end{figure}

However, regarding nonholonomic mobile robots, there does not exist any control input along the direction of $v_y$. Hence, only $v_x$ and $\omega$ can be used to make the robot ``follow" the dynamic vector field $\tilde{\bm{\Gamma}}_{d}^{\mathcal{B}}$. To this end, we introduce an additional rotation in $\omega$, which is a negative feedback relevant to the angle between the orientation of the robot and the direction of $\tilde{\bm{\Gamma}}_{d}^{\mathcal{B}}$. Specifically, as shown in Fig.~\ref{fig_vehicle_attitude}, the orientation of the robot is along the $X_{\mathcal{B}}$-axis of $\bm{\mathcal{F}}_{\mathcal{B}}$, but the integral curve of the vector field moves along the direction of $\tilde{\bm{\Gamma}}_{d}^{\mathcal{B}}$. The angle between these two directions which is denoted by $\delta$ can be given by
\begin{equation}\label{eq_angle_delta}
  \delta=-{\rm atan}\frac{\tilde{\Gamma}_{d}^{\mathcal{B}y}}{\tilde{\Gamma}_{d}^{\mathcal{B}x}},
\end{equation}
where the negative sign means $\delta$ rotates along a counter direction with respect to the attitude angle $\theta$. Note that for nonholonomic mobile robots, the linear velocity $v_x$ has the same direction as the orientation. Therefore, in order to make the robot ``follow" the vector field, the orientation of the robot should be rotate to the direction of $\tilde{\bm{\Gamma}}_{d}^{\mathcal{B}}$. Then, we can construct an additional angular velocity, which is an negative feedback of $\delta$, so as to rotate the robot orientation to the direction of $\tilde{\bm{\Gamma}}_{d}^{\mathcal{B}}$. Thus, based on (\ref{eq_fully_controller}), the motion planning controller for nonholonomic mobile robots is designed to be
\begin{subequations}\label{eq_nonholo_controller}
  \begin{align}
    v_x &= k_v\tilde{\Gamma}_{d}^{\mathcal{B}x}, \\
    \omega &= -k_{\omega}\theta+k_a{\rm atan}\frac{\tilde{\Gamma}_{d}^{\mathcal{B}y}}{\tilde{\Gamma}_{d}^{\mathcal{B}x}},
  \end{align}
\end{subequations}
where $k_v,k_{\omega},k_a$ are all positive scalars.

\section{Dynamic Vector Fields with Obstacle Avoidance}\label{sec_ob_avoid}

This section focuses on the motion planning problem involving obstacle avoidance. Instead of using a repulsive vector field directly, we design a circular vector field around the obstacle, and then blend it with the dynamic vector field which is convergent to the desired configuration.

Consider a circular obstacle with radius $r_o$ located at $\bm{p}_o=[x_o\ \ y_o]^{\rm T}$. Suppose that the obstacle avoidance vector field lies in the following area
\begin{equation*}
  \mathcal{D}=\left\{\bm{q}\in\mathbb{R}^2\  \big| \ r_o<\|\bm{q}-\bm{p}_o\|\leq R_o,\ r_o<R_o<R_s \right\},
\end{equation*}
where $R_s$ is the radius of the sensing region $\mathcal{S}_i$. Once the robot enters the area $\mathcal{D}$, we define a repulsive vector
\begin{equation}\label{eq_Gamma_r_ob_avoid}
  \bm{\Gamma}_r=\bm{p}-\bm{p}_o,
\end{equation}
which points from the obstacle to the nonholonomic mobile robot. Then, the distance between obstacle and robot can be obtained by $d=\|\bm{\Gamma}_r\|$. Let $\bm{\Gamma}_r^{\perp}$ denote the a vector satisfying the following two conditions
\begin{align}\label{eq_Gamma_r_perp_ob_avoid}
  \langle \bm{\Gamma}_r^{\perp},\bm{\Gamma}_r \rangle = 0, \quad
  \langle \bm{\Gamma}_r^{\perp},\bm{v} \rangle > 0,
\end{align}
where $\bm{v}$ is the velocity vector in the body-fixed frame $\bm{\mathcal{F}}_{\mathcal{B}}$ of the nonholonomic mobile robot. Intuitively, the vector $\bm{\Gamma}_r^{\perp}$ is perpendicular to $\bm{\Gamma}_r$ and projected positively onto the velocity direction $\bm{v}$. According to such a definition, we can derive that the vector $\bm{\Gamma}_r^{\perp}$ always points to a collision-free direction with respect to the obstacle, since $\bm{\Gamma}_r^{\perp}$ will be tangent to a certain circle located at $\bm{p}$ with radius $r$ ($r_o<r\leq R_o$). Thus, all the vectors $\bm{\Gamma}_r^{\perp}$ in the area $\mathcal{D}$ can constitute a circular vector field surrounding the obstacle.

Nevertheless, it is fairly unreasonable to construct the obstacle avoidance vector field only by $\bm{\Gamma}_r^{\perp}$, since there must exist some situations where the nonholonomic mobile robot will never collide with the obstacle although moving in the area $\mathcal{D}$. In such cases, it is unnecessary to construct obstacle avoidance vector field for the robot with the help of $\bm{\Gamma}_r^{\perp}$. Otherwise, the results would become rather conservative. Therefore, in the following, we will design the obstacle avoidance vector fields in the area $\mathcal{D}$, denoted by $\bm{\Gamma}_o$, according to different situations of the robot with respect to the obstacle. The detailed construction of $\bm{\Gamma}_o$ are provided below.

\begin{figure}[htp]
  \centering
  \includegraphics[width=0.3\textwidth,trim=0 0 0 0,clip]{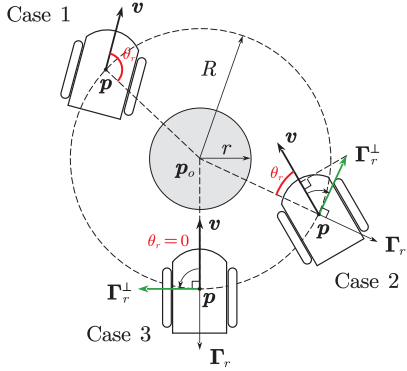}
  \caption{Obstacle avoidance vector fields. Cases~1,2,3 depict three different situations when the robot is approaching the obstacle.}
  \label{fig_vehicle_ob_avoid}
\end{figure}

Assume that the robot has entered the area $\mathcal{D}$, that is, the distance $d$ satisfying $r_o<d\leq R_o$. Let $\theta_r$ represent the angle between the velocity direction $\bm{v}$ and the line segment connecting robot and obstacle, as shown in Fig.~\ref{fig_vehicle_ob_avoid}, which can be computed by
\begin{align*}
  \theta_r &= \arccos\frac{\langle\bm{v},-\bm{\Gamma}_r\rangle}{\|\bm{v}\| \cdot \|\bm{\Gamma}_r\|}
\end{align*}
Hence, according to the value of $\theta_r$, three different cases can be proposed as follows.

\begin{enumerate}
  \item If $\theta_r\geq\frac{\pi}{2}$, as shown in Case~1 of Fig.~\ref{fig_vehicle_ob_avoid}, the robot does not have the risk of colliding with the obstacle. Then, in this case, the obstacle avoidance vector field can absolutely be chosen as the convergent dynamic vector field in the free space, i.e., $\bm{\Gamma}_o=\tilde{\bm{\Gamma}}_d$.
  \item If $0<\theta_r<\frac{\pi}{2}$, as shown in Case~2 of Fig.~\ref{fig_vehicle_ob_avoid}, it is possible that the robot will collide with the obstacle in a future time. Thus, we set the obstacle avoidance vector field to be $\bm{\Gamma}_r^{\perp}$, i.e., $\bm{\Gamma}_o=\bm{\Gamma}_r^{\perp}$, which is defined in (\ref{eq_Gamma_r_perp_ob_avoid}). Then, the robot will turn to the direction that is perpendicular to the radius of the obstacle so as to achieve the obstacle avoidance.
  \item Except for above two cases, the rest one is $\theta_r=0$, as shown in Case~3 of Fig.~\ref{fig_vehicle_ob_avoid}. Generally, in such a case, it is referred to as the deadlock of a robot, which corresponds to a local minima of an artificial potential. Once $\theta_r=0$, it can be derived that $\langle \bm{\Gamma}_r^{\perp},\bm{v} \rangle = 0$, indicating the vector $\bm{\Gamma}_r^{\perp}$ is perpendicular to the velocity $\bm{v}$, so that we cannot define the obstacle avoidance vector field by checking whether the projection of $\bm{\Gamma}_r^{\perp}$ onto $\bm{v}$ is positive, as given in (\ref{eq_Gamma_r_perp_ob_avoid}). Actually, the reason why we choose the $\bm{\Gamma}_r^{\perp}$ positively projected onto $\bm{v}$ is that such a vector can make the robot turn a smaller attitude angle for obstacle avoidance than the negatively projected one. However, when $\theta_r=0$, the rotation angle from $\bm{v}$ to $\bm{\Gamma}_r^{\perp}$ is always $\frac{\pi}{2}$, either clockwise or anticlockwise. Thus, we can directly choose one of these two rotation directions to define the obstacle avoidance vector field $\bm{\Gamma}_o$, and in this paper, $\bm{\Gamma}_o$ is defined by rotating $\bm{\Gamma}_r$ clockwise through $\frac{\pi}{2}$, that is, $\bm{\Gamma}_o=\bm{R}_{\frac{\pi}{2}}^{\rm T}\bm{\Gamma}_r$, where $\bm{R}_{\frac{\pi}{2}}$ is rotation matrix given in (\ref{eq_rotation_matrix}) with $\theta=\frac{\pi}{2}$.
\end{enumerate}

To summarize, based on above definitions under different situations, the obstacle avoidance vector field can be given by
\begin{equation}\label{eq_ob_avoid_VF}
  \bm{\Gamma}_o=\left\{
  \begin{aligned}
    &\bm{R}_{\frac{\pi}{2}}^{\rm T}\bm{\Gamma}_r,\quad\quad \theta_r=0, \\
    &\bm{\Gamma}_r^{\perp},\quad\quad 0<\theta_r<\frac{\pi}{2}, \\
    &\tilde{\bm{\Gamma}}_d,\quad\quad\quad\ \theta_r\geq\frac{\pi}{2}.
  \end{aligned}
  \right.
\end{equation}
Therefore, we can present the dynamic vector fields with obstacle avoidance in $\mathbb{R}^2$ as follows
\begin{equation}\label{eq_VF_all_V1}
  \bm{\Gamma}_P=\left\{
  \begin{aligned}
    &\tilde{\bm{\Gamma}}_d, \qquad d>R_o, \\
    &\bm{\Gamma}_o, \quad r_o<d\leq R_o.
  \end{aligned}
  \right.
\end{equation}
Note that the vector field given in (\ref{eq_VF_all_V1}) would be discontinuous at $d=R_o$. To make the vector field continuous, we introduce the following transition function
\begin{equation}\label{eq_switch_func}
  \varsigma=\left\{
  \begin{aligned}
    0,\qquad\qquad   &\qquad d<R_o,\\
    \frac{1}{2}\sin\left(\frac{d-R_o}{\epsilon}\pi-\frac{\pi}{2}\right)+\frac{1}{2},  &\quad R_o\leq d\leq R_o+\epsilon, \\
    1,\qquad\qquad   &\qquad d>R_o+\epsilon,
  \end{aligned}
  \right.
\end{equation}
where $\epsilon$ is a small positive constant. It is obvious that the transition function $\varsigma$ continuously varies from $0$ to $1$ as the distance $d$ varies from $R_o+\epsilon$ to $R_o$. Therefore, the vector fields proposed in (\ref{eq_VF_all_V1}) can be revised to be a continuous form, that is
\begin{equation}\label{eq_VF_all_V2}
  \bm{\Gamma}_P=\varsigma\tilde{\bm{\Gamma}}_d+(1-\varsigma)\bm{\Gamma}_o.
\end{equation}
We summarize above results into the following theorem.

\begin{theorem}\label{theo_DVF_ob_avoid}
  Let $(x_d,y_d,\theta_d)$ denote an arbitrarily-specified final state. The dynamic vector field $\bm{\Gamma}_P$ proposed in (\ref{eq_VF_all_V2}) asymptotically converges to $(x,y,\theta)=(x_d,y_d,\theta_d)$ and avoids the collision with the obstacle meanwhile.
\end{theorem}

\begin{IEEEproof}
  The convergence of $\tilde{\bm{\Gamma}}_d$ has been proved by Theorem~\ref{theo_DVF}, so that we merely need to prove that $\bm{\Gamma}_o$ will not cause collision of the robot with the obstacle. According to (\ref{eq_Gamma_r_ob_avoid})-(\ref{eq_ob_avoid_VF}), the vector field $\bm{\Gamma}_o$ can be rewritten as
  \begin{equation}\label{eq_Gamma_o_proof}
    \bm{\Gamma}_o=\bm{R}_{\pm\frac{\pi}{2}}(\bm{p}-\bm{p}_o),
  \end{equation}
  where $\bm{R}_{\pm\frac{\pi}{2}}$ represents the rotation matrix with $\theta=\pm\frac{\pi}{2}$. Define the following distance function
  \begin{equation}
    f=\| \bm{p}-\bm{p}_o\|^2-r^2,
  \end{equation}
  then there holds $f>0$ for $\forall\bm{p}\in\mathcal{D}$. To ensure obstacle avoidance, the distance function $f$ should be guaranteed always positive. Assume that once the robot enters the obstacle avoidance region $\mathcal{D}$, the inital value of $f$ is a constant $c_0>0$. Taking the time derivative along the vector field $\bm{\Gamma}_o$, we have
  \begin{equation}
    \frac{\rm d}{{\rm d}t}f=2(\bm{p}-\bm{p}_o)^{\rm T}\bm{\Gamma}_o=2(\bm{p}-\bm{p}_o)^{\rm T}\bm{R}_{\pm\frac{\pi}{2}}(\bm{p}-\bm{p}_o)=0,
  \end{equation}
  which demonstrates that the distance function will maintain $f=c_0>0$ in the future time, indicating that integral curve of the vector field $\bm{\Gamma}_o$ will maintain a constant distance with respect to the obstacle, thus achieving obstacle avoidance by a circular motion.
\end{IEEEproof}

Although Theorem~\ref{theo_DVF_ob_avoid} is derived based on only one obstacle, it can be simply extended to the case of multiple obstacles. Regarding $M$ obstacles, the dynamic vector field is designed to be
\begin{equation}\label{eq_VF_all_V3}
  \bm{\Gamma}_P=\prod_{i=1}^M\varsigma_i\tilde{\bm{\Gamma}}_d+\sum_{i=1}^M(1-\varsigma_i)\bm{\Gamma}_{oi},
\end{equation}
where $\tilde{\bm{\Gamma}}_d$ is the convergent dynamic vector field, $\bm{\Gamma}_{oi}$ is the obstacle avoidance vector field around the $i$th obstacle, $\varsigma_i$ is the transition function regarding the $i$th obstacle.

Having obtained the vector field in (\ref{eq_VF_all_V3}), it would not be difficult to design the controller for the nonholonomic mobile robot. Similar to (\ref{eq_DVF_in_Fb}), we transform the vector field $\bm{\Gamma}_{P}$ into the body-fixed frame $\bm{\mathcal{F}}_{\mathcal{B}}$ by
\begin{equation}\label{eq_DVF_ob_avoid_in_Fb}
  \bm{\Gamma}_{P}^{\mathcal{B}}=\bm{R}^{\rm T}\bm{\Gamma}_P\triangleq
  \begin{bmatrix}
    \Gamma_{P}^{\mathcal{B}x} \\
    \Gamma_{P}^{\mathcal{B}y}
  \end{bmatrix}.
\end{equation}
Then, inspired by (\ref{eq_nonholo_controller}), the controller can be given by
\begin{subequations}\label{eq_nonholo_controller_ob_avoid}
  \begin{align}
    v_x &= k_v\Gamma_{P}^{\mathcal{B}x}, \\
    \omega &= -k_{\omega}\prod_{i=1}^M\varsigma_i\theta+k_a{\rm atan}\frac{\Gamma_{P}^{\mathcal{B}y}}{\Gamma_{P}^{\mathcal{B}x}},
  \end{align}
\end{subequations}
where $k_v,k_{\omega},k_a$ are all positive scalars.

\section{Dynamic Vector Fields with Collision Avoidance Among Robots}\label{sec_co_avoid}

In this section, we consider the problem of collision avoidance among multiple nonholonomic mobile robots during the motion planning. For simplicity, the collision avoidance between two robots is taken into account at first.

\begin{figure}[htp]
  \centering
  \subfigure[$\bm{v}_i$ and $\bm{v}_j$ lying in different sides of the line segment $\overline{\bm{p}_i\bm{p}_j}$]{
    \label{fig_co_avoid_org_1}
    \includegraphics[width=0.2\textwidth,trim=0 0 0 0,clip]{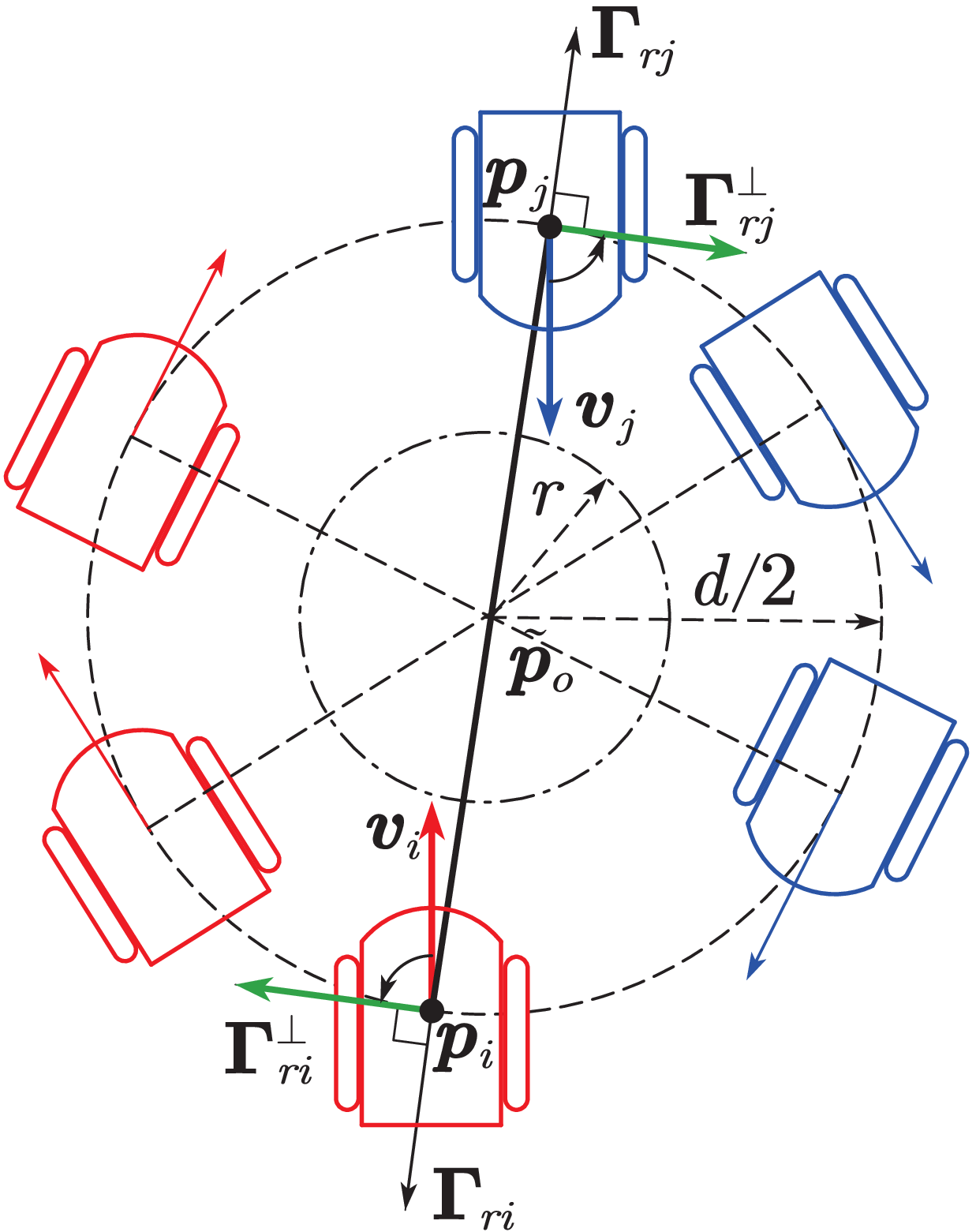}}\quad
  \subfigure[$\bm{v}_i$ and $\bm{v}_j$ lying in the same side of the line segment $\overline{\bm{p}_i\bm{p}_j}$]{
    \label{fig_co_avoid_org_2}
    \includegraphics[width=0.2\textwidth,trim=0 0 0 0,clip]{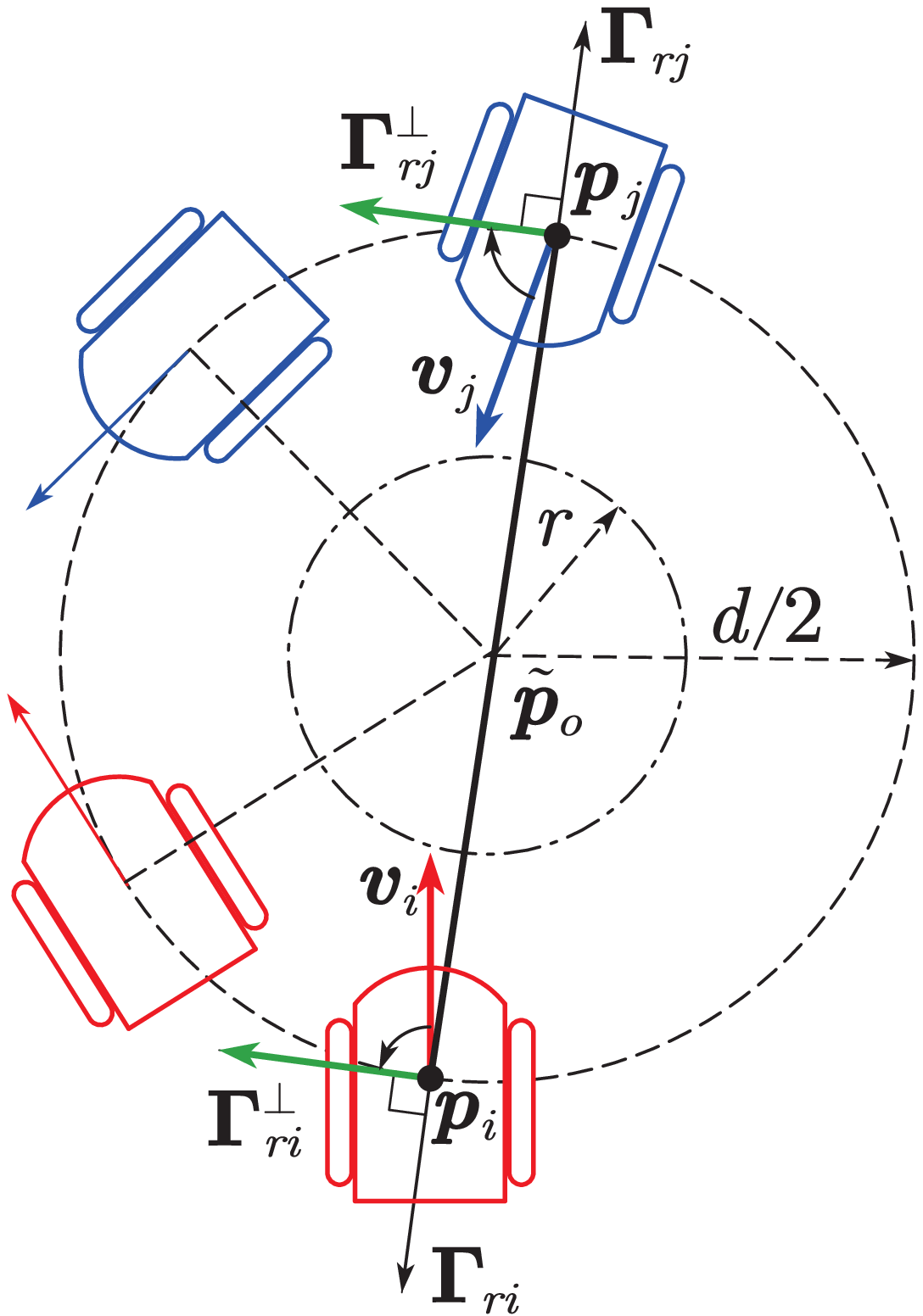}}
  \caption{collision avoidance between two robots in different situations}
  \label{fig_co_avoid_org}
\end{figure}

Motivated by the circular vector field presented in the obstacle avoidance, intuitively, we can introduce a virtual obstacle between two robots so that they are able to avoid each other by avoiding the virtual obstacle. To be more specific, as shown in Fig.~\ref{fig_co_avoid_org_1}, regarding two robots positioned at $\bm{p}_i$ and $\bm{p}_j$, when there is a potential collision risk in their sensing ranges, a virtual obstacle is set on the line segment $\overline{\bm{p}_i\bm{p}_j}$. According to the obstacle avoidance vector field in Section~\ref{sec_ob_avoid}, the robots will turn to the direction of $\bm{\Gamma}_{ri}^{\perp}$ and $\bm{\Gamma}_{rj}^{\perp}$, respectively, as the green vectors in Fig.~\ref{fig_co_avoid_org_1}, which are projected positively along $\bm{v}_i$ and $\bm{v}_j$, respectively. Then, two robots will follow the vectors $\bm{\Gamma}_{ri}^{\perp}$ and $\bm{\Gamma}_{rj}^{\perp}$ to avoid the virtual obstacle clockwise. In this way, the collision avoidance between two robots can be realized completely.

Nevertheless, in Fig.~\ref{fig_co_avoid_org_1}, it should be noted that the velocities $\bm{v}_i$ and $\bm{v}_j$ lie in the different sides of the line segment $\overline{\bm{p}_i\bm{p}_j}$. Once $\bm{v}_i$ and $\bm{v}_j$ lie in the same side of $\overline{\bm{p}_i\bm{p}_j}$, as given in Fig.~\ref{fig_co_avoid_org_2}, the obstacle avoidance vectors $\bm{\Gamma}_{ri}^{\perp}$ and $\bm{\Gamma}_{rj}^{\perp}$ which are projected positively along $\bm{v}_i$ and $\bm{v}_j$ will point to the opposite rotation directions around the virtual obstacle, i.e., clockwise and anticlockwise, respectively. Then, two mobile robots will turn to such directions and cause the collision eventually. Therefore, the obstacle avoidance vector field cannot be extended to collision avoidance straightforwardly by introducing an virtual obstacle between two robots.

\begin{figure}[htp]
  \centering
  \includegraphics[width=0.22\textwidth,trim=0 0 0 0,clip]{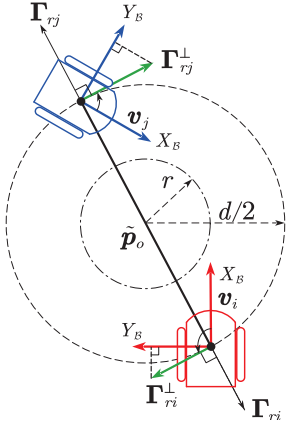}
  \caption{Collision avoidance of two nonholonomic mobile robots by vectors $\bm{\Gamma}_{ri}^{\perp}$ and $\bm{\Gamma}_{rj}^{\perp}$}
  \label{fig_vehicle_co_avoid}
\end{figure}

Based on abovementioned analysis, the key point of the collision avoidance is how to define the direction of vectors $\bm{\Gamma}_{ri}^{\perp}$ and $\bm{\Gamma}_{rj}^{\perp}$ so as to make them both clockwise or both anticlockwise. To solve this problem, we still consider two robots located at $\bm{p}_i$ and $\bm{p}_j$, and suppose that they have been into each other's sensing range, i.e., $d_{ij}=\|\bm{p}_i-\bm{p}_j\|<R_s$. In addition, we assume that the distance threshold for starting collision avoidance denoted by $R_c$ satisfies $R_c<R_s$. That is to say, as shown in Fig.~\ref{fig_vehicle_co_avoid}, when $d_{ij}\leq 2R_c$, two robots are supposed to have a potential risk of collision, and the vector fields should be converted from the mode of target navigation to collision avoidance.

Following the idea of virtual obstacle, we define a circular obstacle with radius $r_{ij}$, whose position vector is given by
\begin{equation}\label{eq_tilde_p_o}
  \tilde{\bm{p}}_o=\frac{1}{2}(\bm{p}_i+\bm{p}_j).
\end{equation}
Note that $2r_{ij}$ can be regarded as the minimum safe distance for no-collision. In other words, two robots will collide with each other if $d_{ij}\leq2r_{ij}$. Thus, the collision avoidance vector field will lie in the following area
\begin{equation*}
  \mathcal{D}_c=\left\{\bm{q}\in\mathbb{R}^2\  \big| \ r_{ij}<\|\bm{q}-\tilde{\bm{p}}_o\|\leq R_c,\ r_{ij}<R_c<R_s \right\},
\end{equation*}
Let us take the robot $i$ for example. Similar to the obstacle avoidance, the repulsive vector $\bm{\Gamma}_{ri}$ can be defined by
\begin{equation}\label{eq_Gamma_r_co_avoid}
  \bm{\Gamma}_{ri}=\bm{p}_i-\tilde{\bm{p}}_o.
\end{equation}
Subsequently, we define the vector $\bm{\Gamma}_{ri}^{\perp}$ perpendicular to $\bm{\Gamma}_{ri}$ as follows
\begin{align}\label{eq_Gamma_r_perp_co_avoid}
  \langle \bm{\Gamma}_{ri}^{\perp},\bm{\Gamma}_{ri} \rangle = 0, \quad
  \langle \bm{\Gamma}_{ri}^{\perp},\bm{R}_{\frac{\pi}{2}}\bm{v}_i \rangle > 0,
\end{align}
where $\bm{R}_{\frac{\pi}{2}}$ is rotation matrix in (\ref{eq_rotation_matrix}) with $\theta=\frac{\pi}{2}$. Note that $\bm{R}_{\frac{\pi}{2}}$ rotates anticlockwise the velocity direction $\bm{v}_i$ by $\frac{\pi}{2}$, then the vector $\bm{R}_{\frac{\pi}{2}}\bm{v}_i$ points exactly to the $Y_{\mathcal{B}}$-axis of the body-fixed frame $\mathcal{F}_{\mathcal{B}}$. Thus, different from (\ref{eq_Gamma_r_perp_ob_avoid}) in the obstacle avoidance, $\bm{\Gamma}_{ri}^{\perp}$ in (\ref{eq_Gamma_r_perp_co_avoid}) becomes the vector which is projected positively onto $Y_{\mathcal{B}}$-axis rather than $X_{\mathcal{B}}$-axis (i.e., the direction of $\bm{v}_i$), as the green vectors illustrated in Fig.~\ref{fig_vehicle_co_avoid}. Due to the fact that for each robot the $Y_{\mathcal{B}}$-axis is always obtained by rotating anticlockwise $X_{\mathcal{B}}$-axis through $\frac{\pi}{2}$, all the robots will rotate anticlockwise to follow the vector field $\bm{\Gamma}_{ri}^{\perp}$ once having a collision risk. Therefore, the robots will move along the anticlockwise rotation direction to accomplish the collision avoidance.

According to $\bm{\Gamma}_{ri}$ given in (\ref{eq_Gamma_r_co_avoid}), for robot $i$, the dynamic vector field with collision avoidance in $\mathbb{R}^2$ can be proposed as follows
\begin{equation}\label{DVF_Gamma_Qi}
  \bm{\Gamma}_{Qi}=\varsigma\tilde{\bm{\Gamma}}_d+(1-\varsigma)\bm{\Gamma}_{ri}^{\perp},
\end{equation}
where the transition function $\varsigma$ is obtained from (\ref{eq_switch_func}) by replacing $R_o$ with $R_c$. Above results are summarized in the theorem below.

\begin{theorem}\label{theo_DVF_co_avoid}
  Let $(x_{di},y_{di},\theta_{di})$ denote an arbitrarily-specified final state for the robot $i$. The dynamic vector field $\bm{\Gamma}_{Qi}$ proposed in (\ref{DVF_Gamma_Qi}) asymptotically converges to $(x,y,\theta)=(x_{di},y_{di},\theta_{di})$ and avoids the collision with other robots.
\end{theorem}

\begin{IEEEproof}
  Similar to the proof of Theorem~\ref{theo_DVF_ob_avoid}, we need to prove that $\bm{\Gamma}_{Qi}$ will not cause collisions between the robots. Define the following distance function
  \begin{equation}
    f_{ij}= \|\bm{p}_i-\bm{p}_j\|^2 - (2r_{ij})^2,
  \end{equation}
  which satisfies $f_{ij}>0$ for $\forall\bm{p}_i,\bm{p}_j\in\mathcal{D}_c$. Taking the time derivative along the vector fields $\bm{\Gamma}_{ri}^{\perp}$ and $\bm{\Gamma}_{rj}^{\perp}$, we have
  \begin{align}\label{eq_dot_f_ij}
    \frac{\rm d}{{\rm d}t}f_c &= 2(\bm{p}_i-\bm{p}_j)^{\rm T}(\bm{\Gamma}_{ri}^{\perp}-\bm{\Gamma}_{rj}^{\perp}) \nonumber \\
    &= 2(\bm{p}_i-\bm{p}_j)^{\rm T}\bm{R}_{\frac{\pi}{2}}^{\rm T}(\bm{p}_i-\tilde{\bm{p}}_o-\bm{p}_j+\tilde{\bm{p}}_o) \nonumber \\
    &=0,
  \end{align}
   which implying that the distance between robot $i$ and robot $j$ will keep constant in the area $\mathcal{D}_c$. Note that the distance function $f_{ij}$ satisfies $f_{ij}>0$ at the initial time for collision avoidance, then there will still hold $f_{ij}>0$ in the future time. Hence, the robot $i$ will not collide with robot $j$ in the movement.
\end{IEEEproof}

Based on the vector field $\bm{\Gamma}$ designed in (\ref{DVF_Gamma_Qi}), we can present the control inputs $v_i$ and $\omega_i$ further. Similar to aforementioned sections, the controller is still proposed with the aid of the vector field given in the body-fixed frame $\mathcal{F}_{\mathcal{B}}$.
Let $\bm{\Gamma}_{Qi}^{\mathcal{B}}$ denote the formulation of $\bm{\Gamma}_{Qi}$ expressed into the body-fixed frame $\bm{\mathcal{F}}_{\mathcal{B}}$, which can be obtained by
\begin{equation}\label{eq_DVF_co_avoid_in_Fb}
  \bm{\Gamma}_{Qi}^{\mathcal{B}}=\bm{R}^{\rm T}\bm{\Gamma}_{Qi}\triangleq
  \begin{bmatrix}
    \Gamma_{Qi}^{\mathcal{B}x} \\
    \Gamma_{Qi}^{\mathcal{B}y}
  \end{bmatrix}.
\end{equation}
Nonetheless, there are two facts worth noting in the design of $v_{xi}$ and $\omega_i$, which are different from the controllers in above sections. One fact is that the linear velocities of the robot $i$ and robot $j$ should be equivalent, when these two robots move around the virtual obstacle for collision avoidance. Actually, the definitions in (\ref{eq_tilde_p_o}) and (\ref{eq_Gamma_r_perp_co_avoid}) implicitly guarantees that $\bm{\Gamma}_{ri}^{\perp}$ and $\bm{\Gamma}_{rj}^{\perp}$ have the same amplitudes, so that $\frac{\rm d}{{\rm d}t}f_c=0$ in (\ref{eq_dot_f_ij}) holds. Therefore, once the robots enter the collision avoidance field, we set the linear velocities to be a common constant denoted by $v_c$, which can be decided according to the range of speed for real robots. The other fact is that the angle between the directions of $\bm{v}_i$ and $\Gamma_{ri}^{\perp}$, similar to the angle $\delta$ defined in (\ref{eq_angle_delta}), belongs to $[-\pi,\pi]$ instead of $[-\frac{\pi}{2},\frac{\pi}{2}]$. This is because the vector $\Gamma_{ri}^{\perp}$ is not ensured for positive projection onto $X_{\mathcal{B}}$-axis, but onto $Y_{\mathcal{B}}$-axis. Then, the angle from the direction of $\bm{v}_i$ to $\Gamma_{ri}^{\perp}$ should be decided by
\begin{equation}\label{eq_angle_delta_2}
  \delta=-{\rm atan2}(\Gamma_{Qi}^{\mathcal{B}y},\Gamma_{Qi}^{\mathcal{B}x})
\end{equation}
Therefore, based on above two facts, the controller can be designed as
\begin{subequations}\label{eq_nonholo_controller_co_avoid}
  \begin{align}
    v_{xi} &= k_v\varsigma\tilde{\Gamma}_d^{\mathcal{B}x}+(1-\varsigma)v_c, \\
    \omega_i &= -k_{\omega}\varsigma\theta+k_a{\rm atan2}(\Gamma_{Qi}^{\mathcal{B}y},\Gamma_{Qi}^{\mathcal{B}x}),
  \end{align}
\end{subequations}
where $k_v,k_{\omega},k_a$ are all positive scalars.

Having investigated the collision avoidance of two robots, it will be not difficult to handle the collision avoidance problem for multiple nonholonomic mobile robots in the motion planning process. Still following above ideas, a virtual obstacle can be introduced for the multi-robot collision avoidance. As we can see from the analysis for two robots, the crucial problem for such an approach is how to define the position of the virtual obstacle. Regarding two robots, the virtual obstacle is set at the middle point of the line segment connecting two robot positions. Such a point can also be interpreted as the convex combination of the two positions with coefficient $\frac{1}{2}$. Inspired by this fact, we can employ the concept of convex combination to decide the position of the virtual obstacle.

At a certain instant, regarding robot $i$, we assume that there are $M_c$ robots ($M_c<N$) whose distance with the robot $i$ is less than the collision avoidance threshold $R_c$. Let $\mathcal{P}$ denote the set which is constituted by the serial numbers of these $M_c$ robots, so that it follows $d_{ij}=\|\bm{p}_i-\bm{p}_j\|<R_c$ for $\forall j\in\mathcal{P}$. The positions of these $M_c$ robots are denoted by $\bm{p}_1,\bm{p}_2,\cdots,\bm{p}_{M_c}$, then we can define the following convex combination
\begin{equation}
  \tilde{\bm{p}}_o=\frac{1}{M_c}\sum_{i=1}^{M_c}\bm{p}_i.
\end{equation}
Further, we place the virtual obstacle at the convex combination $\tilde{\bm{p}}_o$ and make these robots move round the virtual obstacle, as the case of two robots given in Fig.~\ref{fig_vehicle_co_avoid}, so as to achieve the collision avoidance among these robots. Once the position of the virtual obstacle is decided, the collision avoidance vector field $\bm{\Gamma}_{ri}^{\perp}$ can be easily obtained by (\ref{eq_Gamma_r_co_avoid}) and (\ref{eq_Gamma_r_perp_co_avoid}). In addition, the control inputs can also be straightforwardly derived as (\ref{eq_nonholo_controller_co_avoid}).

\section{Numerical Simulation Examples}\label{sec_sim}

In this section, four numerical simulation examples are provided, i.e., 1) motion planning in an obstacle-free space, 2) motion planning with obstacle avoidance, 3) coordinated motion planning with collision avoidance, 4) coordinated motion planning with both obstacle and collision avoidance.

\begin{table}[htp]
\renewcommand{\arraystretch}{1.5}
\caption{Initial and final states for motion planning in an obstacle-free space}
\label{tab_MP_Initial_Final}
\centering
\begin{tabular}{c|c|c}
\hline
Case No. & \makecell{Initial condition \\ $(x_0,y_0,\theta_0)$} & \makecell{Specified final state \\ $(x_d,y_d,\theta_d)$} \\
\hline
1 & \multirow{6}*{$(0,0,0)$} & $(0,40,0)$ \\
2 & ~ & $(40,40,\frac{\pi}{2})$ \\
3 & ~ & $(40,0,-\frac{\pi}{2})$ \\
4 & ~ & $(40,-40,0)$ \\
5 & ~ & $(-20,-40,-\frac{\pi}{2})$ \\
6 & ~ & $(-40,0,\pi)$ \\
\hline
\end{tabular}
\end{table}

\begin{figure}[htp]
  \centering
  \subfigure[Initial and specified final states of the robot]{
    \label{fig_MP_Initial_Final}
    \includegraphics[width=0.23\textwidth,trim=60 5 70 5,clip]{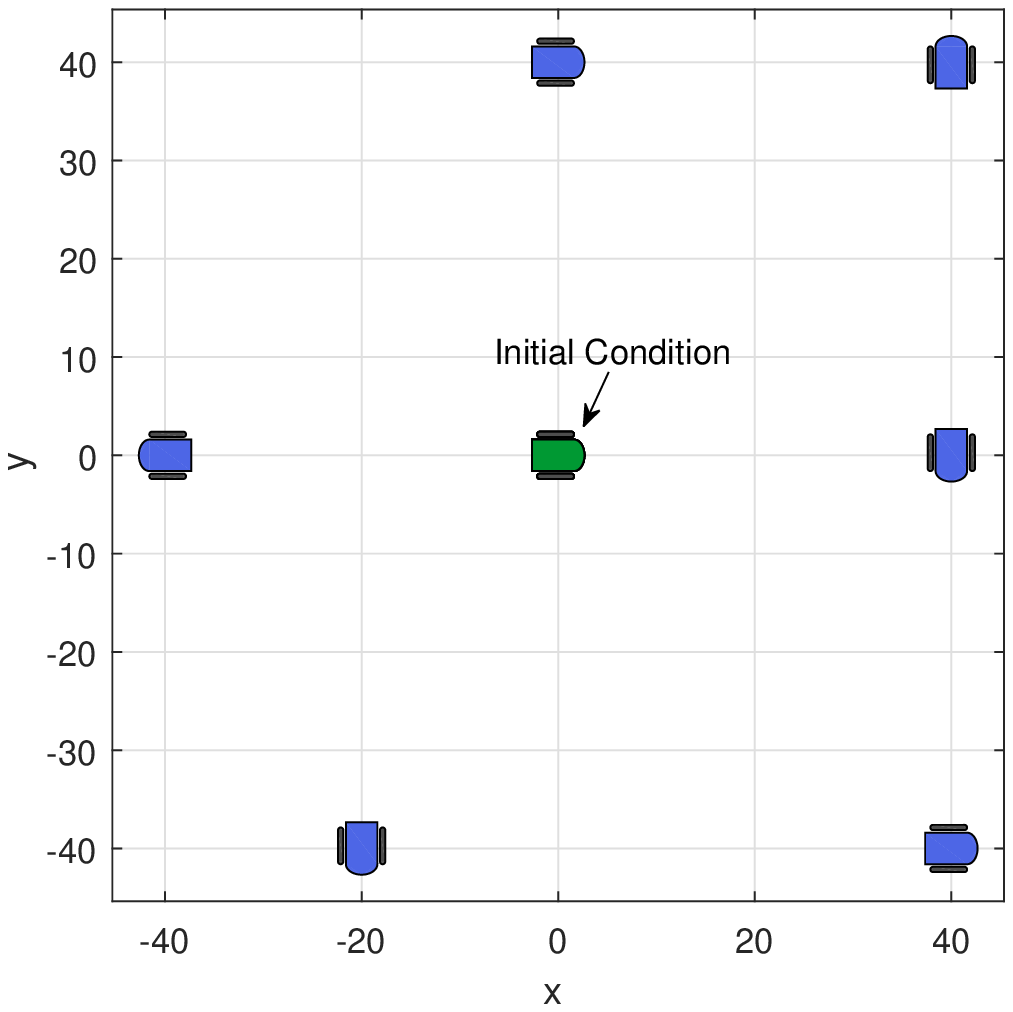}}
  \subfigure[Trajectories of the robot by motion planning]{
    \label{fig_MP_Trajectory}
    \includegraphics[width=0.23\textwidth,trim=60 5 70 5,clip]{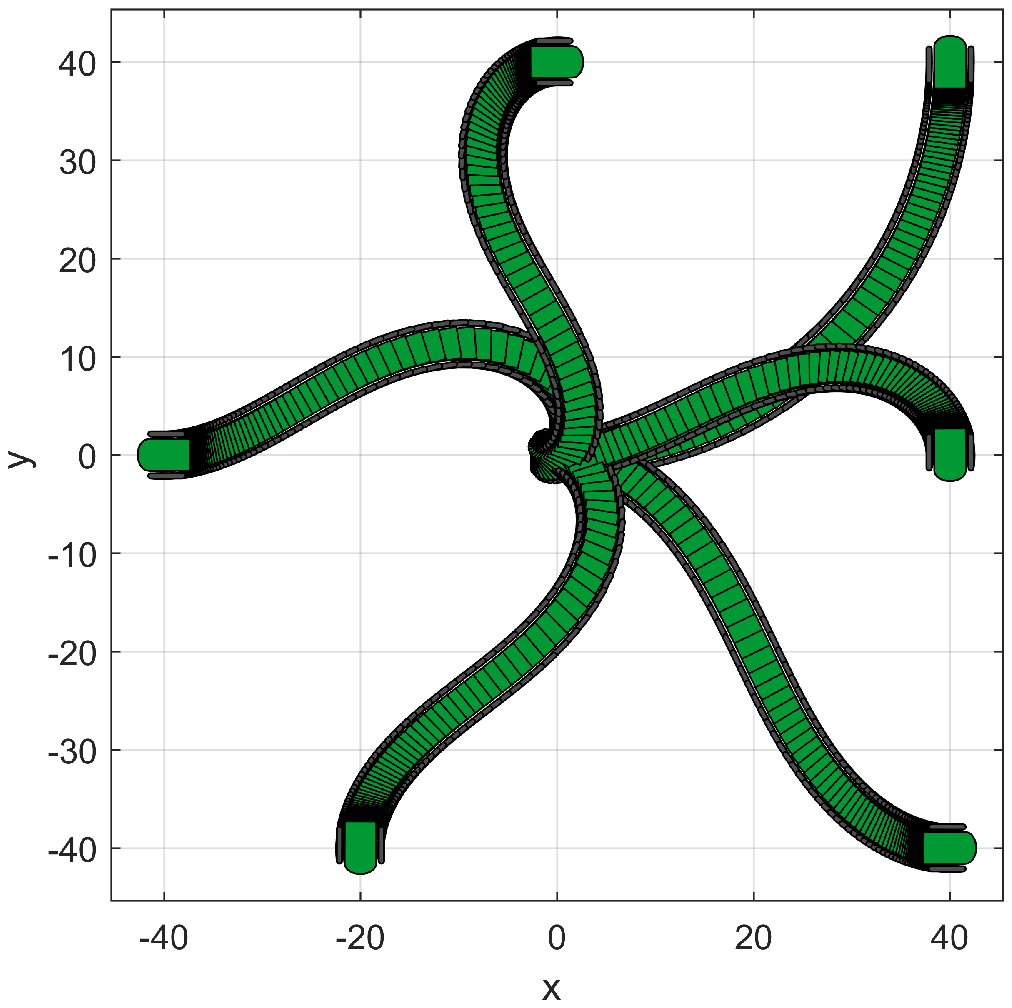}}
  \caption{Simulation results of motion planning in an obstacle-free space}
  \label{fig_MP_Sim}
\end{figure}

\begin{example}[Motion planning in an obstacle-free space]
  The dynamic vector field (\ref{eq_DVF_arbitr}) is employed in this example to verify the effectiveness of motion planner in an obstacle-free space. The initial and final states of the nonholonomic mobile robot are given in Table~\ref{tab_MP_Initial_Final}, where we choose six different final states including positions and orientations. Note that this is an example for one robots under various requirements of final states, instead of coordinated motion planning of multiple robots. Fig.~\ref{fig_MP_Initial_Final} depicts the initial and specified final states of the robot. It should be mentioned that the case of $(x_d,y_d,\theta_d)=(0,40,0)$ is a quite challenging one, because it is in the lateral direction and has a same attitude as the initial state, while the robot cannot move sideways. The simulation results of robot trajectories are provided in Fig.~\ref{fig_MP_Trajectory}, which shows that the robot reaches the specified positions as well as the desired orientations.
\end{example}

\begin{table}[htp]
\renewcommand{\arraystretch}{1.5}
\caption{Initial and final states for motion planning with obstacle avoidance}
\label{tab_OB_Initial_Final}
\centering
\begin{tabular}{c|c|c}
\hline
Case No. & \makecell{Initial condition \\ $(x_0,y_0,\theta_0)$} & \makecell{Specified final state \\ $(x_d,y_d,\theta_d)$} \\
\hline
1 & $(0,30,0)$ & \multirow{3}*{$(0,0,0)$} \\
2 & $(-30,30,\frac{\pi}{2})$ & ~ \\
3 & $(-35,0,\pi)$ & ~ \\
\hline
\end{tabular}
\end{table}

\begin{figure*}[htp]
  \centering
  \subfigure[$t=0$s]{
    \label{fig_ob_Sim1}
    \includegraphics[width=0.235\textwidth,trim=60 5 70 5,clip]{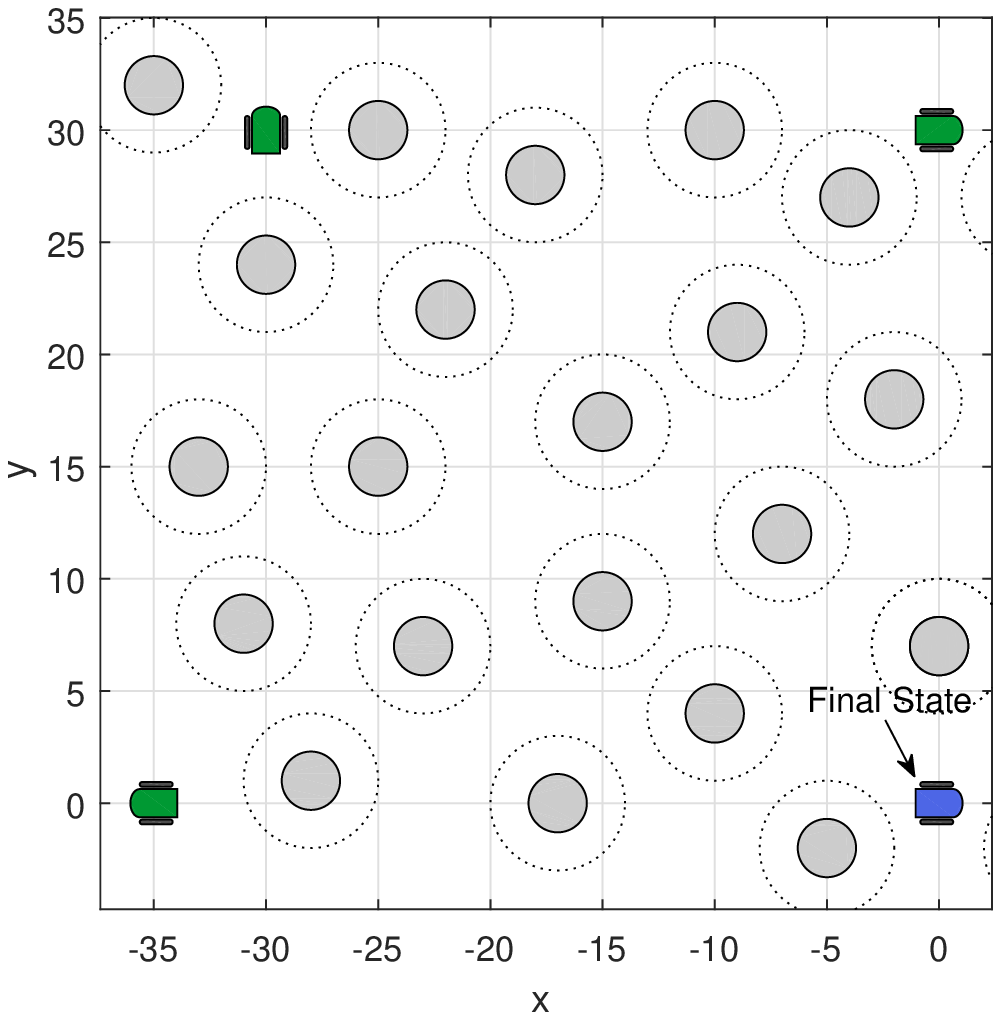}}
  \subfigure[$t=5$s]{
    \label{fig_ob_Sim2}
    \includegraphics[width=0.235\textwidth,trim=60 5 70 5,clip]{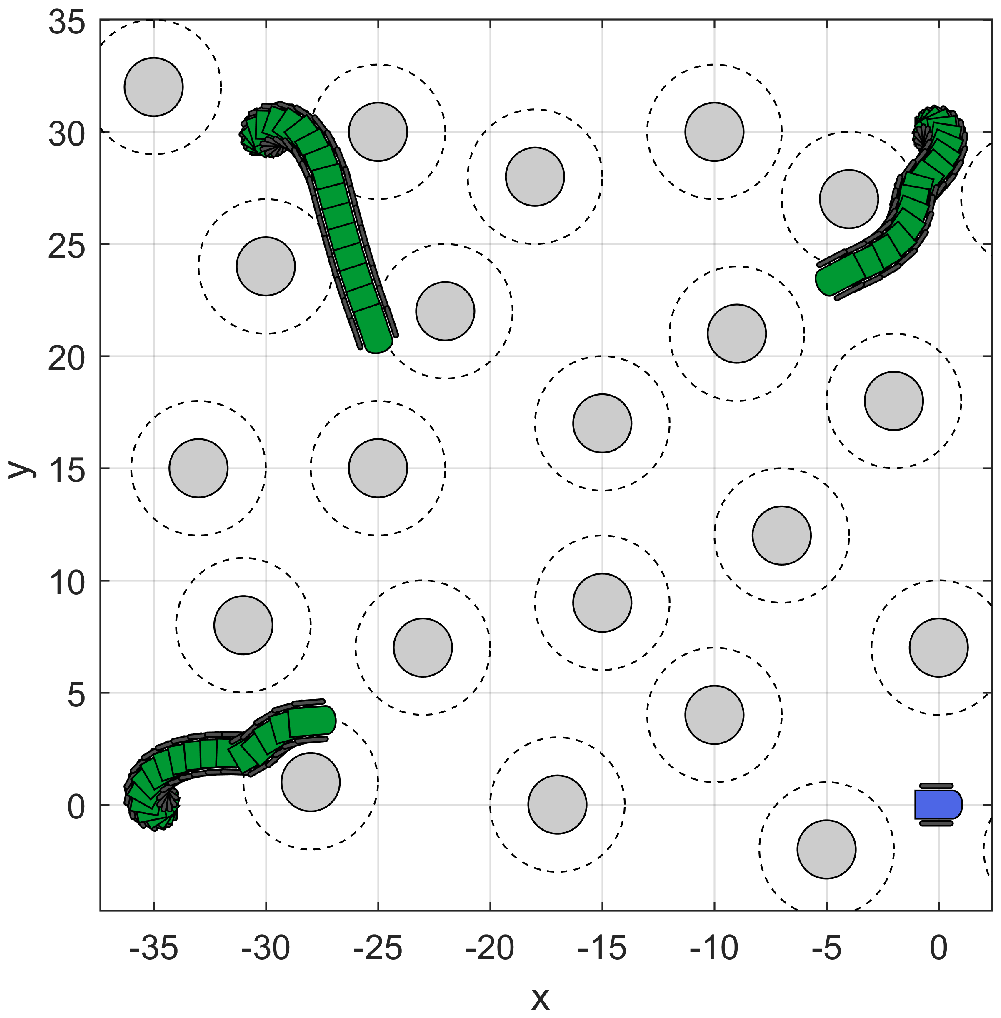}}
  \subfigure[$t=10$s]{
    \label{fig_ob_Sim3}
    \includegraphics[width=0.235\textwidth,trim=60 5 70 5,clip]{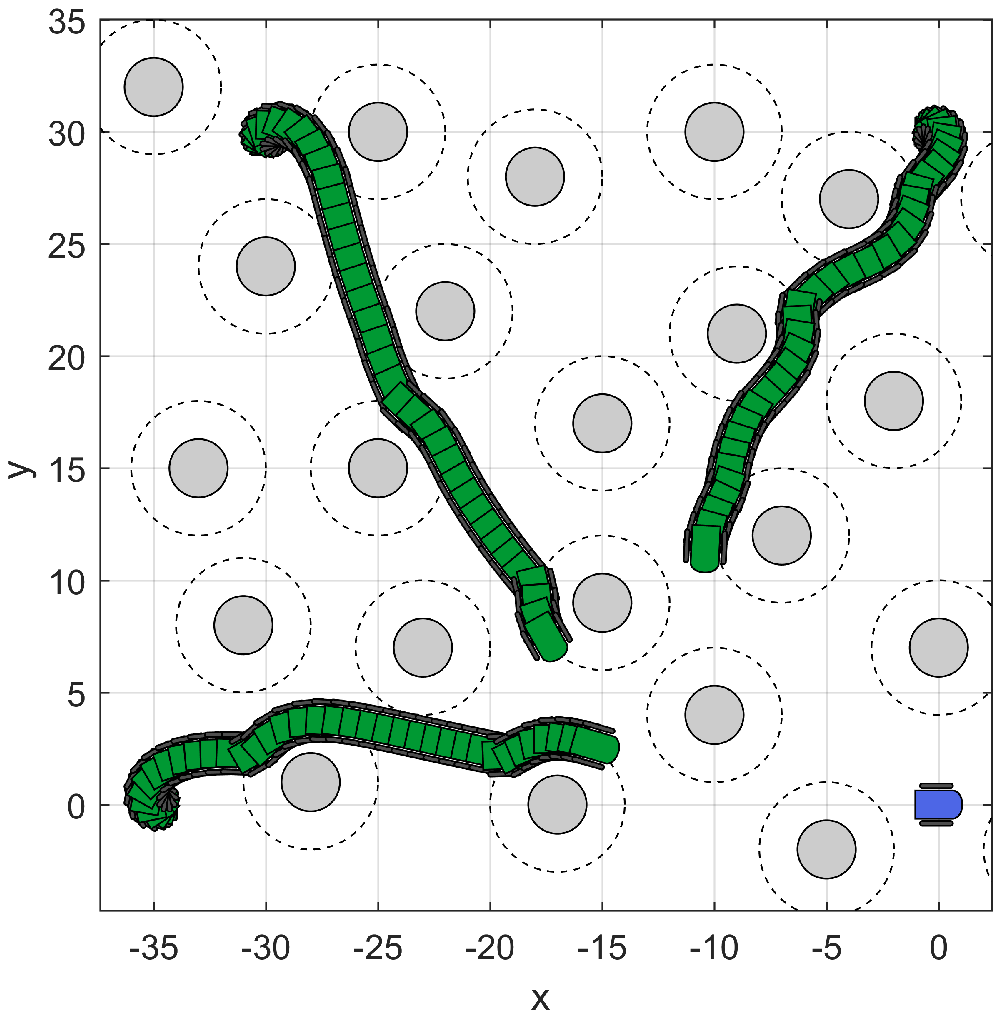}}
  \subfigure[$t=20$s]{
    \label{fig_ob_Sim4}
    \includegraphics[width=0.235\textwidth,trim=60 5 70 5,clip]{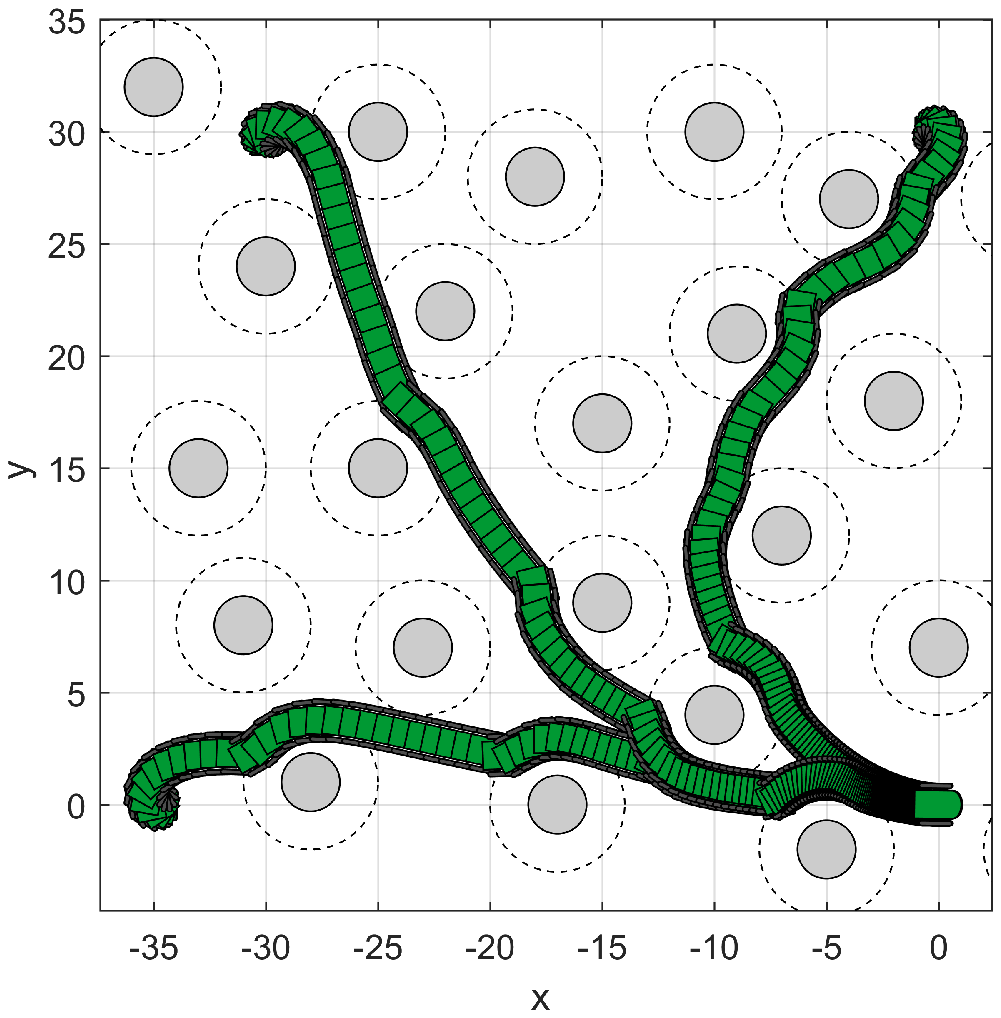}}
  \caption{Simulation results of motion planning in an obstacle environment}
  \label{fig_ob_Sim}
\end{figure*}

\begin{figure*}[htp]
  \centering
  \subfigure[$t=0$s]{
    \label{fig_co_line_Sim1}
    \includegraphics[width=0.235\textwidth,trim=60 5 70 5,clip]{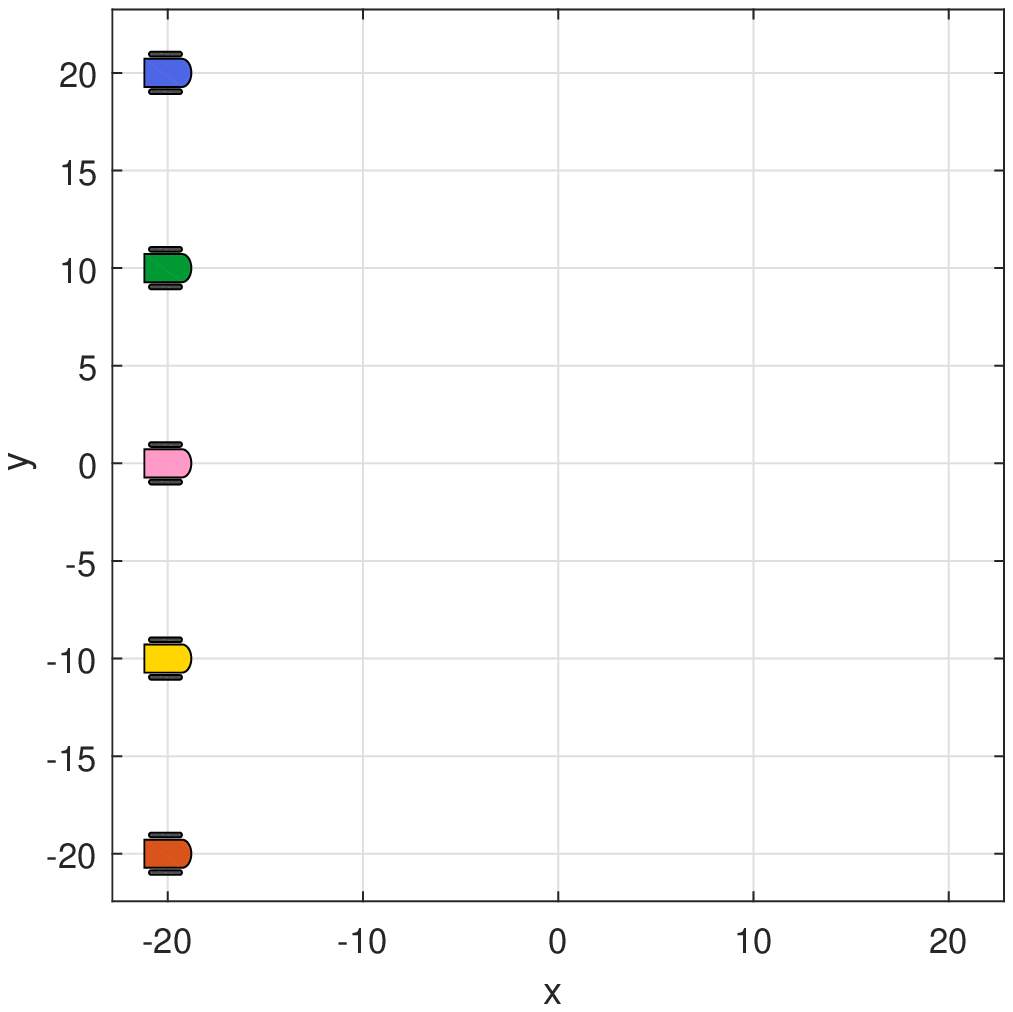}}
  \subfigure[$t=3$s]{
    \label{fig_co_line_Sim2}
    \includegraphics[width=0.235\textwidth,trim=60 5 70 5,clip]{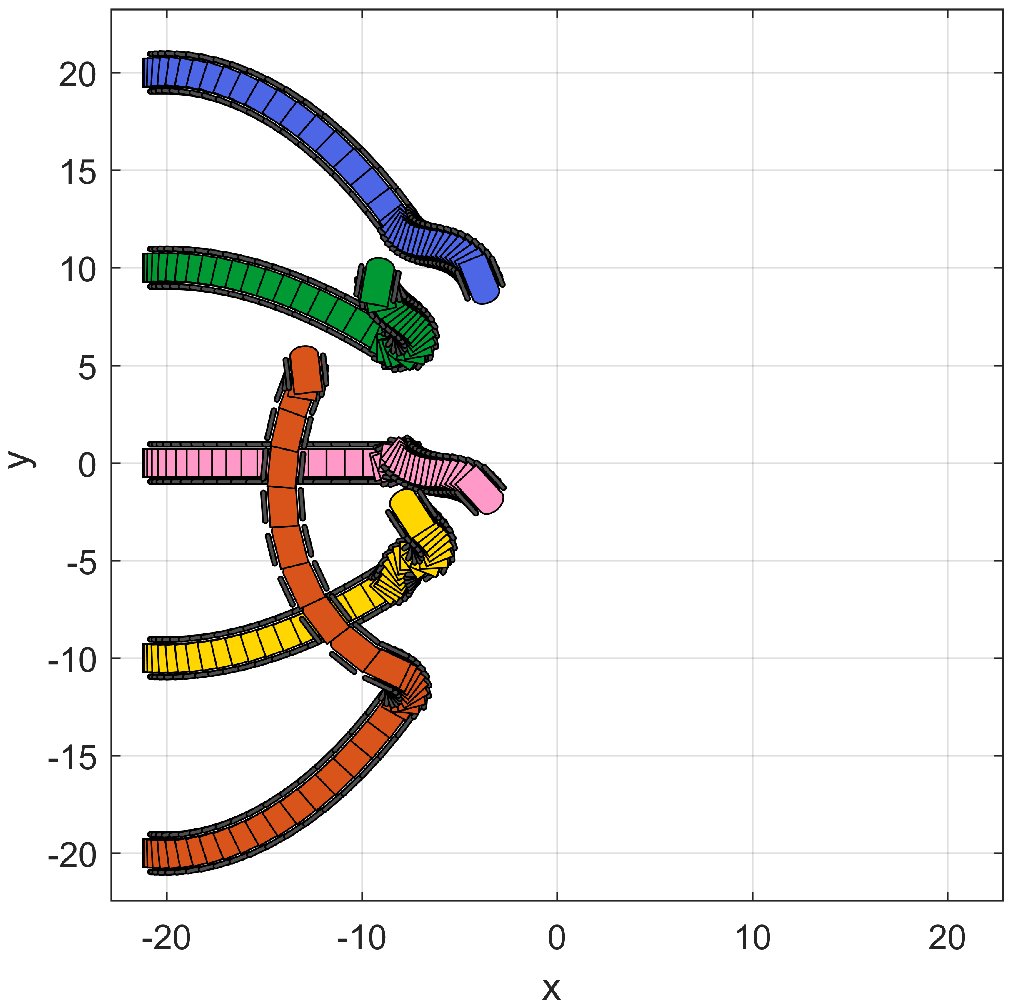}}
  \subfigure[$t=6$s]{
    \label{fig_co_line_Sim3}
    \includegraphics[width=0.235\textwidth,trim=60 5 70 5,clip]{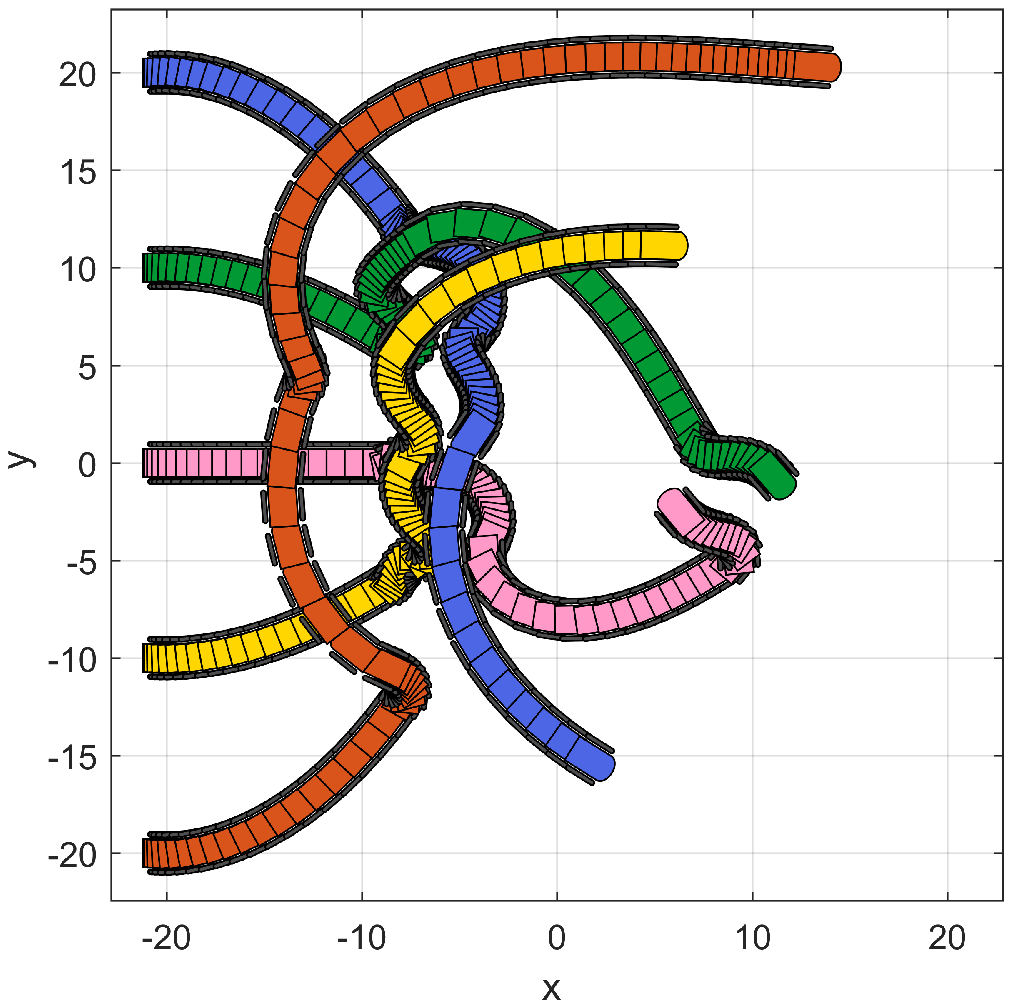}}
  \subfigure[$t=15$s]{
    \label{fig_co_line_Sim4}
    \includegraphics[width=0.235\textwidth,trim=60 5 70 5,clip]{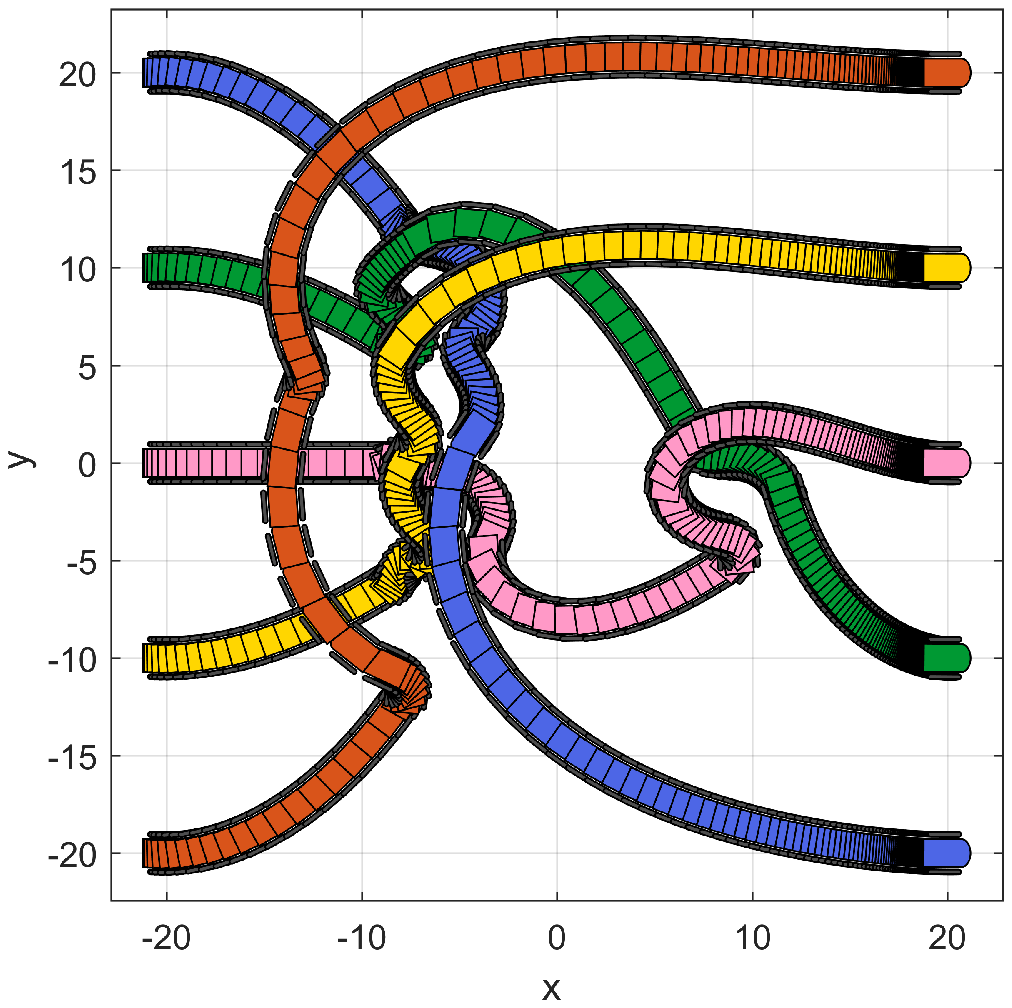}}
  \caption{Simulation results of coordinated motion planning with collision avoidance (Scenario~1)}
  \label{fig_co_line_Sim}
\end{figure*}

\begin{figure*}[htp]
  \centering
  \subfigure[$t=0$s]{
    \label{fig_co_circle_Sim1}
    \includegraphics[width=0.235\textwidth,trim=60 5 70 5,clip]{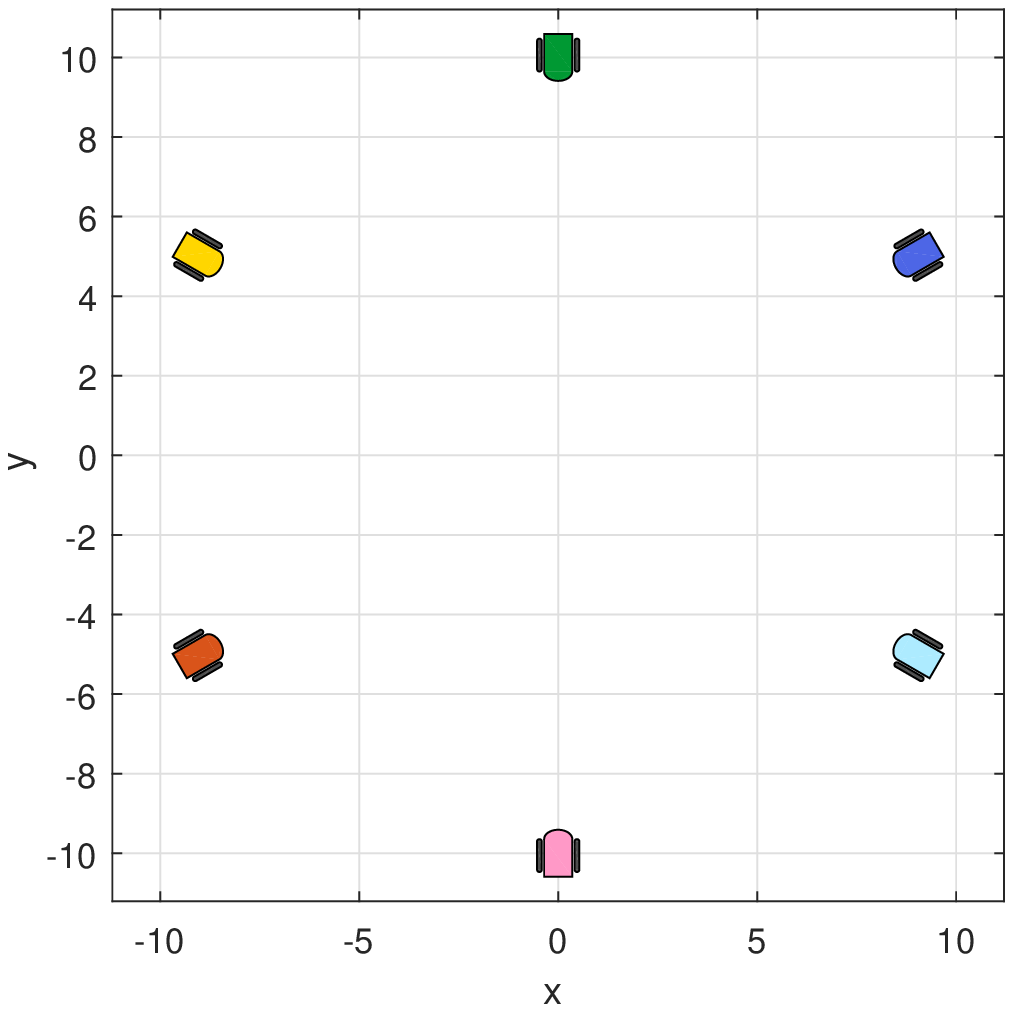}}
  \subfigure[$t=3$s]{
    \label{fig_co_circle_Sim2}
    \includegraphics[width=0.235\textwidth,trim=60 5 70 5,clip]{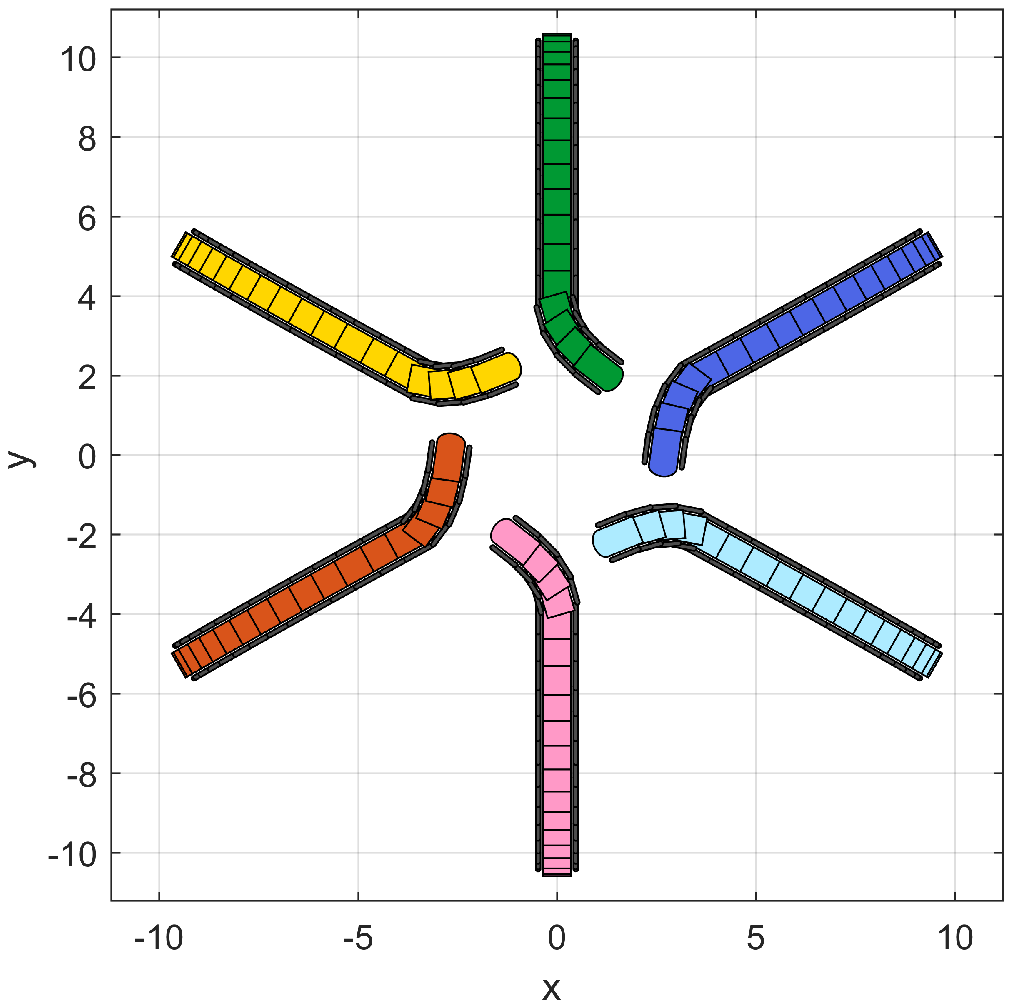}}
  \subfigure[$t=6$s]{
    \label{fig_co_circle_Sim3}
    \includegraphics[width=0.235\textwidth,trim=60 5 70 5,clip]{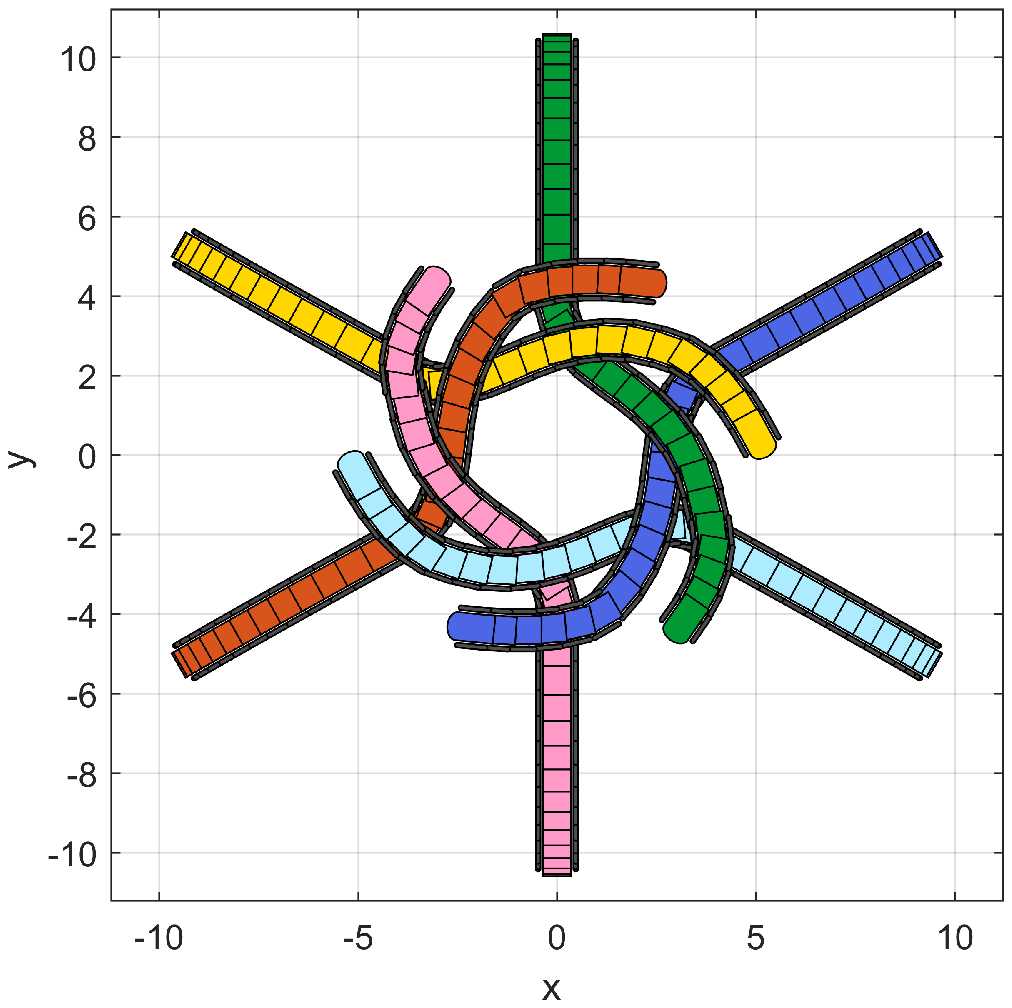}}
  \subfigure[$t=15$s]{
    \label{fig_co_circle_Sim4}
    \includegraphics[width=0.235\textwidth,trim=60 5 70 5,clip]{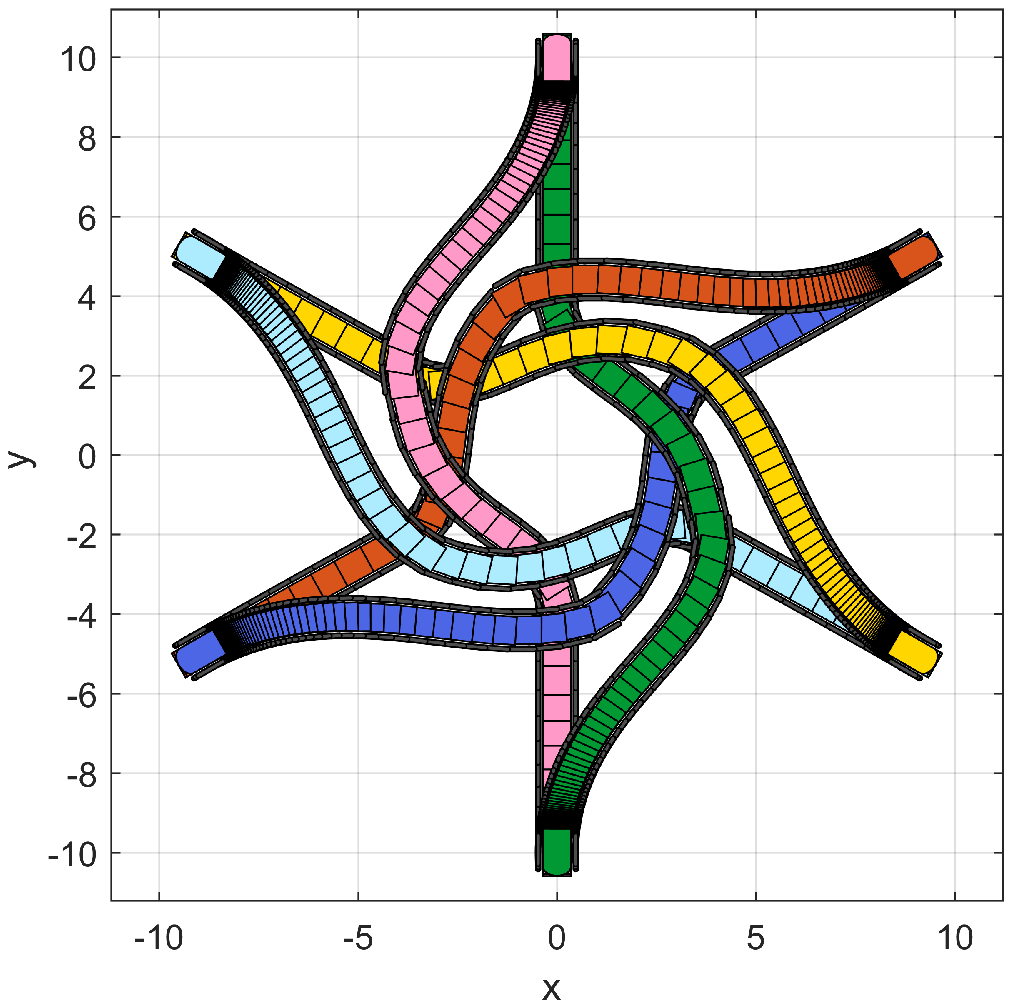}}
  \caption{Simulation results of coordinated motion planning with collision avoidance (Scenario~2)}
  \label{fig_co_circle_Sim}
\end{figure*}

\begin{example}[Motion planning with obstacle avoidance]
  This example considers the motion planning in an obstacle environment, so that the dynamic vector field with obstacle avoidance should be utilized. The initial and final states of one robot is provided in Table~\ref{tab_OB_Initial_Final}, where we choose three different initial conditions. The radius of the obstacle is set to be $r_o=1.5$, while the radius of the region with obstacle-avoidance vector field is $R_o=3$. The simulation time is chosen as $T=20$s, and the trajectories of three different scenarios are given in Fig.~\ref{fig_ob_Sim}. It can be seen that the robot can arrive at the specified final states and avoid the obstacles meanwhile.
\end{example}

\begin{example}[Coordinated motion planning with collision avoidance]
  In this example, we provide two scenarios of coordinated motion planning of multiple robots with collision avoidance. In the first scenario, five robots start from a line formation with parallel orientations, as shown in Fig.~\ref{fig_co_line_Sim1}. These robots are required to reach their own position in another line formation and keep the same orientations as their initial ones. The trajectories of the robots at different time instants are illustrated in Fig.~\ref{fig_co_line_Sim2}-Fig.~\ref{fig_co_line_Sim4}, which indicates that achieve the desired final states and avoid collisions with each other. The second scenario considers six robots on a circle, which are expected to exchange their positions with the opposite one and at the same time maintain the initial orientations at the final instant. Such a scenario cause possibly deadlocks for artificial potential or optimization methods, but the vector field proposed in this paper is applicable to this problem, which can be observed from the simulation results in Fig.~\ref{fig_co_circle_Sim}.
\end{example}

\begin{figure*}[htp]
  \centering
  \subfigure[$t=0$s]{
    \label{fig_ob_co_Sim1}
    \includegraphics[width=0.235\textwidth,trim=60 5 70 5,clip]{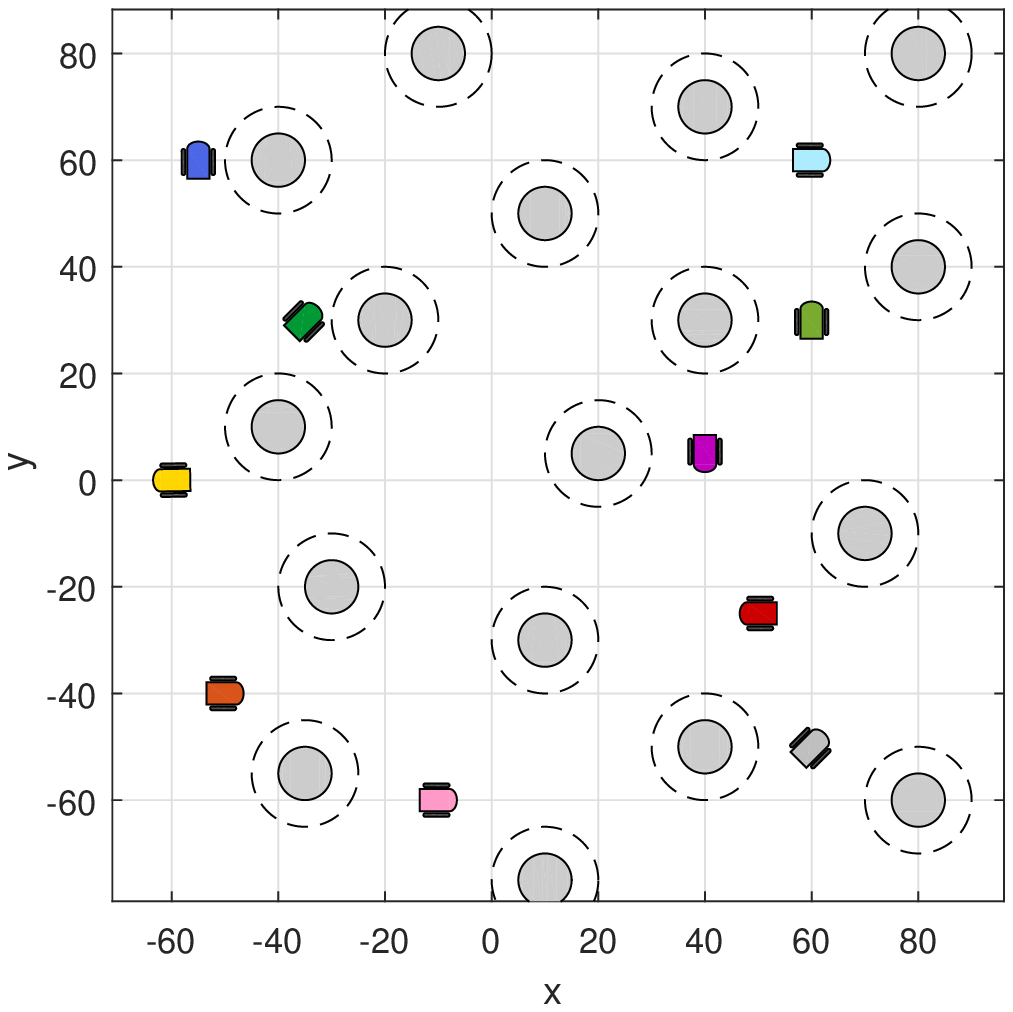}}
  \subfigure[$t=6$s]{
    \label{fig_ob_co_Sim3}
    \includegraphics[width=0.235\textwidth,trim=60 5 70 5,clip]{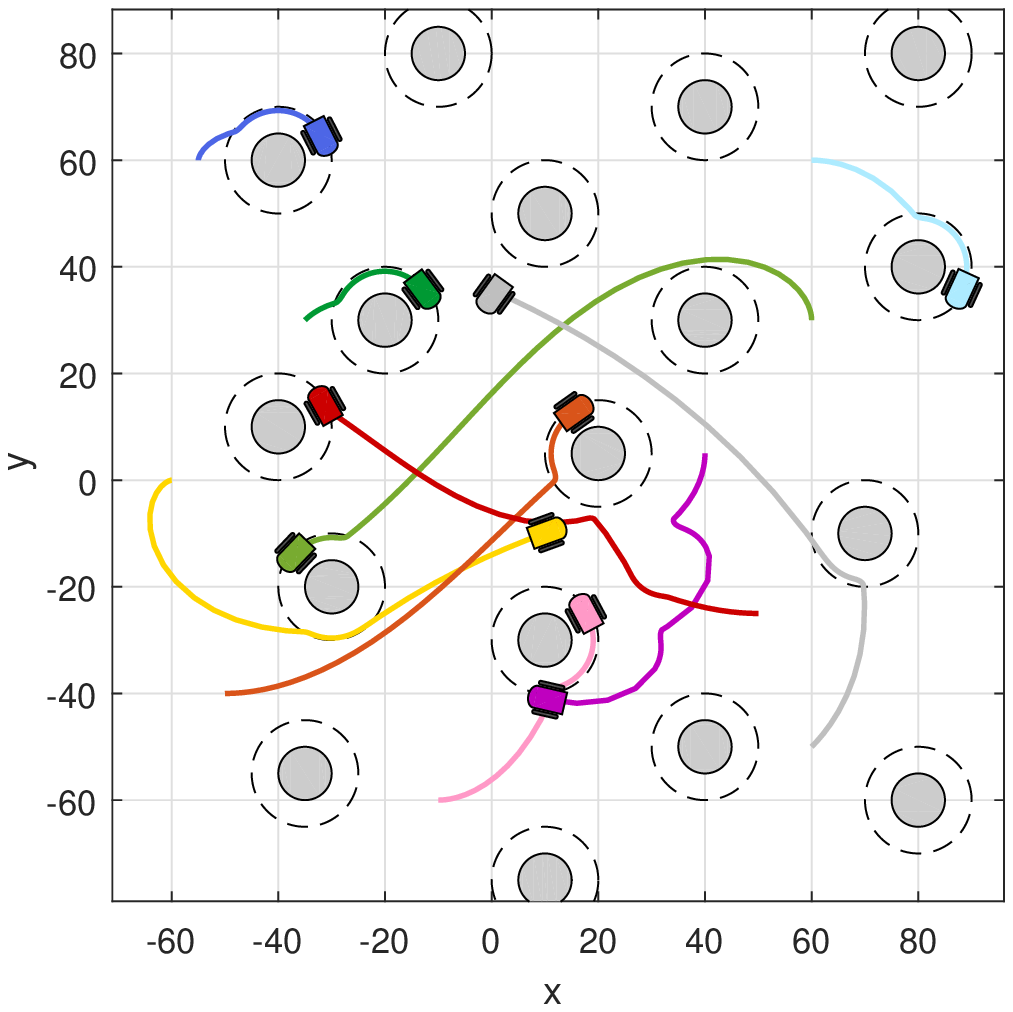}}
  \subfigure[$t=15$s]{
    \label{fig_ob_co_Sim5}
    \includegraphics[width=0.235\textwidth,trim=60 5 70 5,clip]{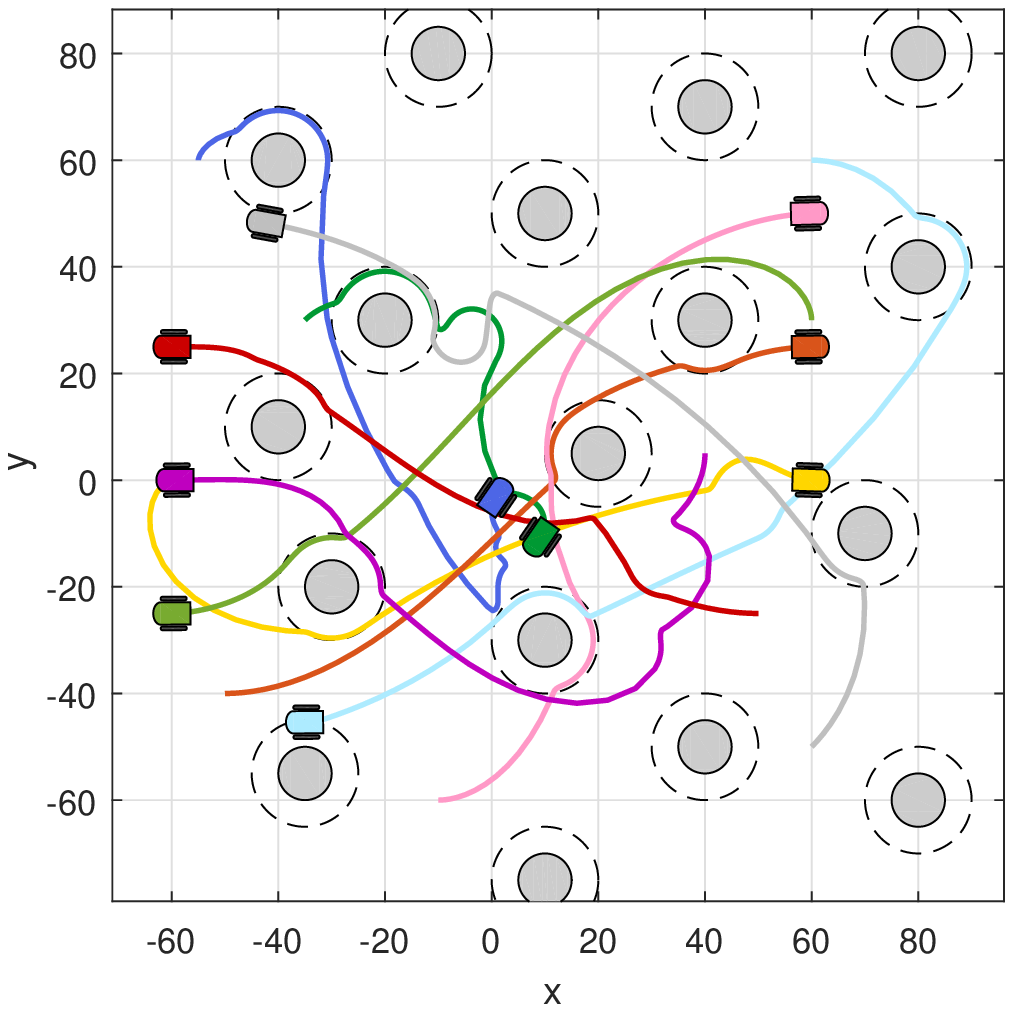}}
  \subfigure[$t=25$s]{
    \label{fig_ob_co_Sim6}
    \includegraphics[width=0.235\textwidth,trim=60 5 70 5,clip]{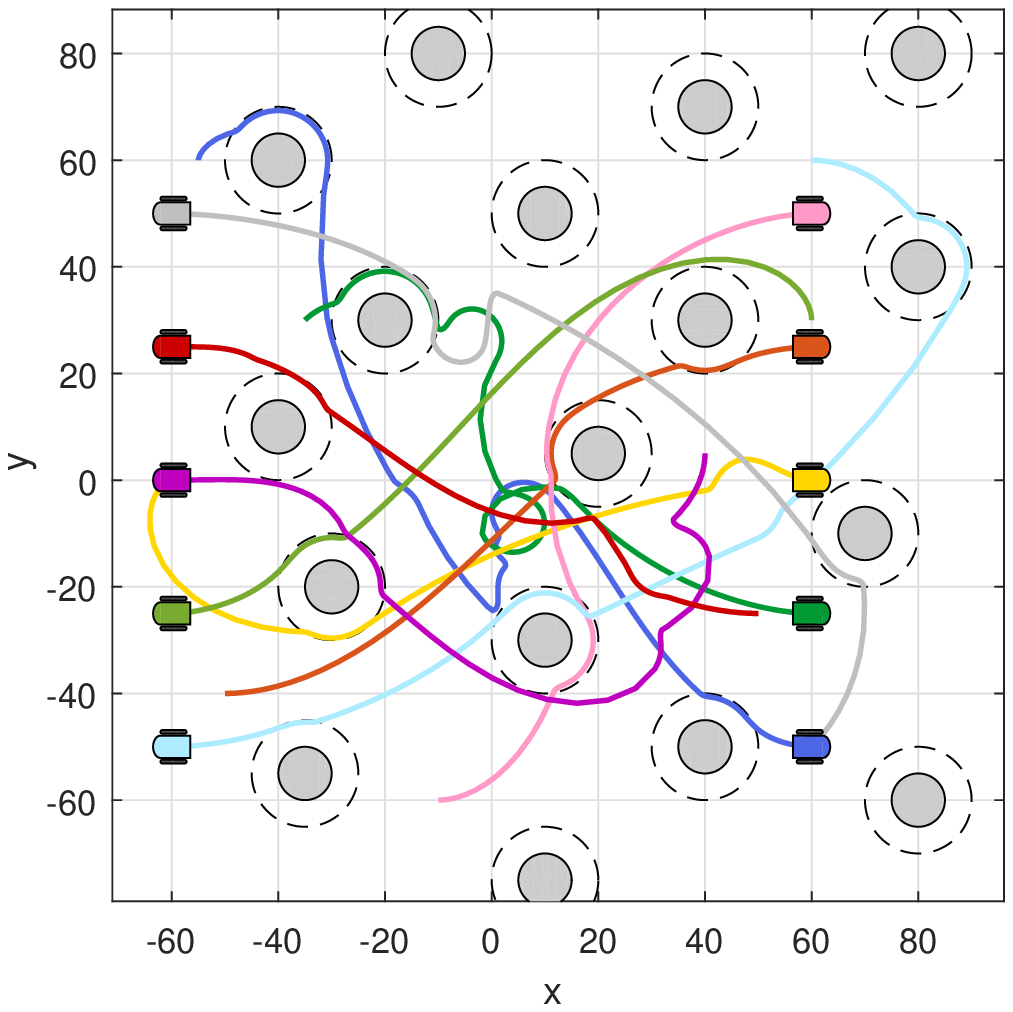}}
  \caption{Simulation results of coordinated motion planning with both of obstacle and collision avoidance}
  \label{fig_ob_co_Sim}
\end{figure*}

\begin{figure}[htp]
  \centering
  \includegraphics[width=0.35\textwidth,trim=20 5 20 5,clip]{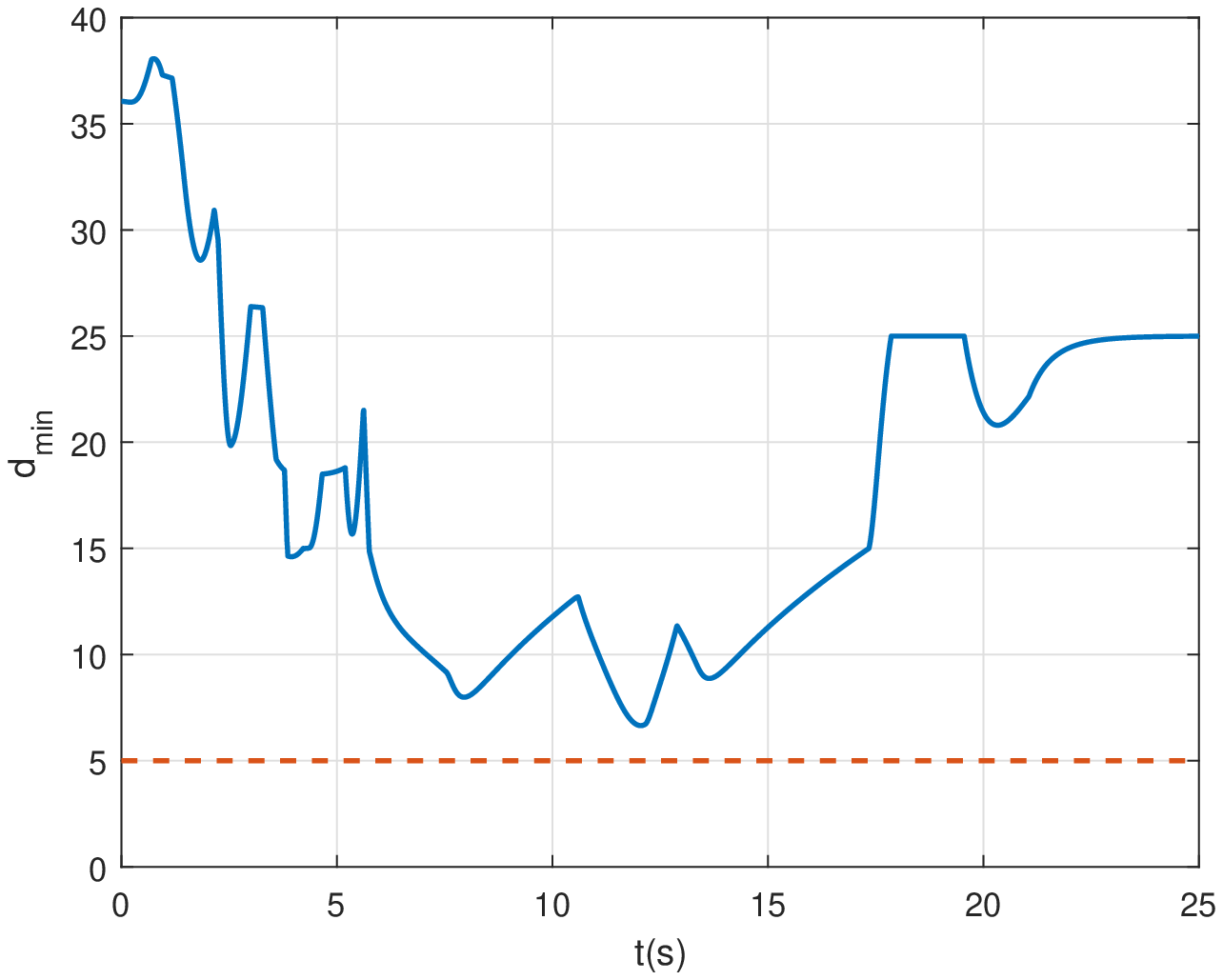}
  \caption{Minimum distance among all pairs of robots at each time instant}
  \label{fig_distance}
\end{figure}

\begin{example}[Coordinated motion planning with both obstacle and collision avoidance]
  The last example handles the multi-robot motion planning in an obstacle environment, so that we have to take into account the obstacle avoidance as well as the collision avoidance with each other. In this example, the number of the robots is $N=10$ and the safe distance between any pair of robots is set to be $r_{ij}=5$. Besides, the radius of the obstacle is chosen to be $r_o=5$, while the radius of the region with obstacle-avoidance vector field is $R_o=10$. The motions of all robots at different time instants are depicted in Fig.~\ref{fig_ob_co_Sim}, in which the final states of the robots are line formations with desired parallel orientations. It can also be viewed from Fig.~\ref{fig_ob_co_Sim} that the trajectories of the robots have no overlaps with obstacles, implying that the mission of obstacle avoidance is realized successfully. The minimum distance among all pairs of robots at each time instant is shown in Fig.~\ref{fig_distance}, which demonstrates that the safe threshold is not violated during the motion period.
\end{example}

\section{Conclusion}\label{sec_con}

This paper has studied the simultaneous position and orientation planning of multiple nonholonomic mobile robots. Such a planning problem takes into account the position and orientation requirements simultaneously, indicating that the robot can reach the goal point with a specified attitude angle. In contrast to the existing open-loop algorithms, we have proposed a novel global feedback motion planning method, namely, a dynamic vector field, under which the directions of velocity vectors over the 2-D plane are decided by both of position and orientation and the nonholonomic constraint can be handled. In addition, by blending with a circular vector field, the dynamic vector field has been extended to the cases of obstacle and collision avoidance. Future works will focus on the vector-field-based motion planning under more realistic occasions, such as input saturations, external disturbances, measurement noises.

\ifCLASSOPTIONcaptionsoff
  \newpage
\fi



\bibliographystyle{IEEEtran}
\bibliography{IEEEabrv,mybibfile}
\end{document}